\documentclass[runningheads]{llncs}
\usepackage{graphicx}
\usepackage{subfigure}
\usepackage{amsmath}
\usepackage{mathabx}
\usepackage{booktabs}
\usepackage{algorithm}
\usepackage{algorithmicx}
\usepackage{algpseudocode}
\usepackage{xcolor}
\usepackage{cleveref}
\usepackage{url}
\usepackage{soul}
\usepackage[misc]{ifsym}

\begin{document}
\title{Automatic Generation of Product Concepts from Positive Examples, with an Application to Music Streaming}
%
%
\author{Kshitij Goyal$^1$ \and
Wannes Meert$^1$ \and
Hendrik Blockeel$^1$ \and
Elia Van Wolputte$^1$ \and
Koen Vanderstraeten$^2$ \and
Wouter Pijpops$^2$ \and
Kurt Jaspers$^2$
}
\authorrunning{K. Goyal et al.}
\titlerunning{Automatic Generation of Product Concepts from Positive Examples}
\institute{KU Leuven, Belgium \and Tunify, Belgium}
\maketitle              
\begin{abstract}
Internet based businesses and products (e.g. e-commerce, music streaming) are becoming more and more sophisticated every day with a lot of focus on improving customer satisfaction.
A core way they achieve this is by providing customers with an easy access to their products by structuring them in catalogues using navigation bars and providing recommendations.
We refer to these catalogues as {\em product concepts}, e.g. product categories on e-commerce websites, public playlists on music streaming platforms.
These {\em product concepts} typically contain products that are linked with each other through some common features (e.g. a playlist of songs by the same artist).
How they are defined in the back-end of the system can be different for different products.
In this work, we represent product concepts using database queries and tackle two learning problems.
First, given sets of products that all belong to the same unknown product concept, we learn a database query that is a representation of this product concept.
Second, we learn product concepts and their corresponding queries when the given sets of products are associated with multiple product concepts.
To achieve these goals, we propose two approaches that combine the concepts of PU learning with Decision Trees and Clustering. 
Our experiments demonstrate, via a simulated setup for a music streaming service, that our approach is effective in solving these problems.

\end{abstract}

\section{Introduction}
Machine learning is used for various applications these days and more and more businesses are looking to use machine learning to improve their products. 
Recent advances in online consumer based businesses provide an opportunity to explore machine learning solutions to challenging problems.
One such problem is generating dynamic product concepts that contain a selection of products that are linked with each other through some common features.
For example, in e-commerce a product category that contains similar items (e.g., `cosmetic items') is a product concept, in music streaming services a public playlist is a product concept.
Typically these kinds of product concepts are available for users to select from.
There are benefits of having such product concepts in your system: 1. they provide a good way to structure the itinerary; 2. a user can use them to navigate the website; 3. they make it easy for users to discover new items.

Due to the lack of transparency behind how these product concepts are created for most services, we can not generalize their creation in the back-end of a system.
However, in our work, we assume that a product concept is associated with a database query, which we term as the concept query, that filters that whole database of items (products, songs etc.) based on certain common features.
This makes these product concepts dynamic in nature: they get automatically updated when new items are added to the database.
This definition is inspired by our use case of a music streaming company called {\bf Tunify}\footnote{\url{https://www.tunify.com/nl-be/language/}}.

Tunify is a music streaming service that provides a predefined selection of playlists to businesses.
Tunify has a database of songs where each song is represented by a fixed set of discrete valued features: {\em mood, popularity} etc.
Tunify also maintains a set of database queries that define useful product concepts (the product concept defined by a query is a set of all songs that are returned when that query is run on the database).
These products concepts are useful for generating playlists for businesses.
A business can select a product concept based on a small description (e.g., 80's Rock) and a playlist based on the selected product concept is generated, the generated playlist is a sample of all the songs associated with the product concept.
The database queries corresponding to product concepts are manually defined by music experts that Tunify employs.
This query creation process has an obvious drawback: it requires a lot of time to fine tune the exact feature-values the query should contain.
This motivates our first problem: can we automatically identify the database query corresponding to a product concept if we are provided with a set of playlists that the target product concept should generate?
We argue that it is easier for an expert to manually create playlists that the target concept should generate compared to manually creating a database query.

Another interesting problem setting is when the provided playlists come from multiple target product concepts.
Can we identify different concepts the playlists are coming from and their corresponding database queries? 
This problem is motivated by the fact that Tunify allows for customers to create their own playlists under one of their subscriptions.
As these playlists come from multiple customers, we expect that: 1. not all of them contain similar songs; 2. there are multiple playlists that contain similar songs.
We want to identify similar playlists and create product concepts based on them.
There are two outcomes of this: either the identified product concepts are missing from Tunify's system, or the identified product concepts are already in the system but the customers didn't use them for whatever reasons.
In the former case, we improve the database with new concepts, and in the later case, Tunify could reach out to the customers that created their own playlists and recommend the already existing concepts.

Even though we motivate these problems based on our use case of Tunify, these problems are generally applicable for any business where such product concepts are used.
Consider the example of an e-commerce company, here the product concepts are the product categories (e.g., {\em cosmetics, menswear}).
The customers interact with the product concept in this case differently from Tunify: in Tunify, a playlist is generated when a customer selects a product concept, but in the case of e-commerce a customer can view all the products associated with a product concept.
In the context of our two learning problems, however, this difference does not have any impact.
For the first learning problem, instead of experts making a playlist of songs, the experts create a set of items.
For the second problem, instead of using playlists of songs created by customers, we can use the items the customers purchased together.

To summarize, we consider two learning problems: 1. learning the product concept query given a collection of itemsets that are associated with it,
2. learning product concepts and their corresponding queries given a collection of itemsets that may or may not be associated with a single product concept.
For the first problem, we use a combination of PU learning (Positive and Unlabelled Learning) \cite{bekker2020learning} techniques with decision tree learning to learn the product concept queries as the rules from the decision tree.
In addition to this, we also study the effect of noise in the provided itemsets on the final query and propose a way to deal with it.
For the second problem, we combine the approach of task one with clustering to identify new product concepts from data generated by customers.

We design and test our experiments on the dataset provided to us by Tunify. 
With a simulated experimental setup, we demonstrate that our approaches are able to learn good quality concept queries with small number of items, even when there is noise in the set, and we are able to effectively identify product concepts using the customer data. 
We additionally show that our proposed algorithm for the first problem is robust to noise in the provided set of items.

The paper is structured in the following way: first we introduce some terminologies and explain the problem statements in section 2, secondly we present our approach in section 3, then we present the experimental results in section 4 before the related works and a discussion in sections 5 and 6 respectively.
Section 7 concludes.

\section{Framework}
In this section, we first give an overview of some concepts from logic and satisfiability which we use in our work before explaining our problem statement.

\subsection{Propositional Logic}
Propositional Logic formulas contain literals which are Boolean formulas, their negation and logical connectives, e.g., $(a \lor (b \land \neg(c)))$.
An assignment $x$ of variables $\{a, b, c\}$ satisfies a formula $\phi$ if $x$ makes the formula $\phi$ True.
Any logical formula can be rewritten in a normal form such as Conjunctive Normal Form (CNF) or Disjunctive Normal Form (DNF).
A CNF formula consists of conjunction of disjunction of literals and a DNF formula contains disjunction of conjunction of literals, where conjunction is the logical `AND' ($\land$) operator and disjunction is the logical `OR' ($\lor$) operator.

\subsection{Problem Statement}
We now formally define our learning problems. 
The dataset of instances is represented by $\mathcal{D}$.
We assume that an instance is represented by a fixed set of discrete valued features $\mathcal{F}$ and takes a single value for each feature.
For a feature $f \in \mathcal{F}$, the discrete set of values $f$ can take is represented by $V(f)$. 

\begin{definition}{\textbf{Product Concept.}}
A product concept is a collection of instances. A product concept is associated with a concept query that defines which instances belong to it.
\label{def:product_concept}
\end{definition}

A product concept can be a union of multiple `sub-concepts'. For example, in e-commerce, a category can have many different sub categories; in Tunify, there are a number of product concepts that combine the music from multiple different product concepts to generate a playlist that contains songs from all the combined product concepts (e.g., `Fitness Center' product concept combines product concepts `Dance Workout' and `Rock Dynamic').
Keeping this in mind, we formally define a concept query as follows:

\begin{definition}{\textbf{Conjunctive Concept Query.}}
Given a set of attributes $F \subseteq \mathcal{F}$ and sets of values $V^f \subseteq V(f)$ for each $f \in F$. A conjunctive concept query $Q$, for an arbitrary input $x \in \mathcal{D}$, is defined as the following rule-based query in a conjunctive normal form:
$$Q: \bigwedge_{f \in F} \bigvee_{v \in V^f} (x_f = v)$$
\label{def:conjunctive_query}
\end{definition}

\begin{definition}{\textbf{Concept Query.}}
A concept query is defined as:
\begin{enumerate}
    \item A conjunctive concept query is a concept query.
    \item A disjunction of two or more conjunctive concept queries is a concept query.
\end{enumerate} 
\label{def:concept_query}
\end{definition}

In the case where a concept query is a disjunction of two or more conjunctive concept queries, each conjunctive concept query and all the items that make it true are said to be associated with a sub-concept of the parent product concept (where the parent product concept is the disjunctive combination of the sub-concepts).
Any item that is associated with any of the sub-concepts is said to belong to the parent product concept.
By definition, a sub-concept is also a product concept.
We will refer to the concept query corresponding to a product concept $C$ as $Q_C$ in the text from now on.
Also, for a given query $Q$, the items from the database $\mathcal{D}$ that are filtered by the query are denoted by $Q(\mathcal{D})$.

We now define our learning problems.
The first problem we consider is the problem of learning the concept query for a target concept given a collection of set of instances that are associated with it.
Note that this problem can be simplified: given that all the provided sets of instances correspond to the same target concept, we can combine them into one set of instances.
This is the standard PU learning setting \cite{bekker2020learning}, where we want to learn just from positive instances.
Formally we define the first learning problem as:

\begin{definition}{\textbf{Learning Problem 1.}}
Given a set of positive instances $S \subset C$, for a target concept $C$, find the corresponding concept query.
\label{def:lp_1}
\end{definition}

For the second learning problem, we want to learn product concepts and their corresponding queries provided a collection of sets of instances. 
The difference from problem \ref{def:lp_1} is that the sets do not necessarily belong to a single target concept.
We do not know how many product concepts are to be learned.
Different sets may belong to the same concept, or to different ones; we do not know which sets belong to the same concept.  
The task is still to learn a concept query for each concept.
Note that problem 2 reduces to problem 1 once we find out which sets belong to the same concept: merging all sets belonging to one concept yields the set of positive instances for that concept.
Formally, the second problem is:

\begin{definition}{\textbf{Learning Problem 2.}}
Given a collection of sets of instances $P_j$, $j=1, \ldots, m$, find a partition of the collection into equivalence classes $K_i$ and corresponding instance sets $S_i = \bigcup_{P \in K_i} P$ such that $S_i \subset C_i$ for each concept $C_i$ to be learned; and find the concept query associated with each concept.
\label{def:lp_2}
\end{definition}

\section{Approach}
We present our approaches for the two learning problems in sections 3.1 and 3.2.

\subsection{Problem 1: Learning Concept Query from a Set of Items}

For the first learning problem, we observe that a concept query is essentially a binary classifier that predicts if a given item belongs to a product concept or not.
From this perspective, since we are only provided with a small number of positive instances, the learning setting is of PU learning (Positive and Unlabelled Learning) \cite{bekker2020learning}.
Following a standard approach in PU learning, we approach this problem by generating the so called {\em reliable negatives} \cite{bekker2020learning} from the full data $\mathcal{D}$, 
which is discussed in more detail later in this section.
Once we have generated the data, we fit a binary classifier that separates the positive instances from the negatives.
Finally, we extract the rule based query from the classifier that can be used as the concept query in the form described in definition \ref{def:concept_query}.
Next subsections provide a more detailed discussion.
The pseudo code is available in algorithm \ref{algo}.

\begin{algorithm}[t]
\caption{ConceptQueryLearner}\label{algo}
\hspace*{\algorithmicindent} \textbf{input:} initial set $\mathbf{S}$, all items $\mathbf{\mathcal{D}}$, discard threshold $\mathbf{d}$
\begin{algorithmic}[1]
\State $Query = \{\}$
\State $N$ = GetReliableNegatives($S$, $\mathcal{D}$)
\State $\mathcal{T}$ = DecisionTree($S$, $N$)
\For{each positive leaf $L$ in $\mathcal{T}$}
    \If{$Size(L) \geq d*Size(S)$}
        \State q = GetQuery(L)
        \State $Query = Query \cup q$
    \EndIf
    \EndFor
\State \Return $Query$
\end{algorithmic}
\label{algo}
\end{algorithm}

\subsubsection{Getting Reliable Negatives:}
Getting reliable negatives is a standard approach for a PU learning problem.
The core assumption behind this is that the data point that are very different from the provided positive instances are likely to be negative examples. 
There are a number of approached proposed for this in the literature \cite{bekker2020learning}.
We employ two approaches for this purpose: First one is a novel probabilistic approach which we call the {\bf likelihood approach} and the second one is the standard {\bf rocchio approach} \cite{li2003learning}.

In the first approach, we calculate the probabilities for each value for each feature based on $S$. For a feature $f \in \mathcal{F}$ and a given value of the feature $v \in V(f)$, the probability is simply calculated as: 
\begin{equation}
P(f = v | S) = \sum_{s \in S}1_{\{s_f = v\}}/|S|    
\label{eq:prob1}
\end{equation}

We assume these probabilities to be the marginal probabilities of the distribution of the target context.
For an item $x \in \mathcal{D} \setminus S$, we calculate the probability that $x$ was sampled from the distribution defined by $S$. Let's assume that item $x$ is defined as $\{color=red, material=m_1, ...\}$, we write the probability as: 
\begin{equation}
P(x) = P(color=red, material=m_1, ...)   
\label{eq:prob2}
\end{equation}
This probability is always smaller than or equal to the marginal probabilities:
\begin{equation}
P(x) \leq P(color=red); P(x) \leq P(material=m_1), ...  
\label{eq:prob3}
\end{equation}

If any of the feature values in $x$ has a marginal probability $0$, this would imply that the $P(x) = 0$ according to equation \ref{eq:prob3}.
Hence, all the items $x \in \mathcal{D} \setminus S$ that have $P(x) = 0$ are chosen to be reliable negatives. Important to note that the set $S$ is a subset of the items in the target concept $C$, and the marginal probabilities based on $S$ are just an approximation of the true marginal. A larger initial set $S$ would lead to better approximation which in turn would lead to better quality of reliable negatives.
We will study the impact of the size of $S$ on the learned context queries in the next section.

The second approach we use is the existing {\em Rocchio} approach \cite{li2003learning} which is based on the rocchio classification \cite{ceri2013introduction}. This method builds prototypes for the labelled ($S$) and the unlabelled data ($\mathcal{D} \setminus S$).
The prototype in our implementation is the centroid of the data points in a binarized version of the dataset $\mathcal{D}$ where each feature value pair is replaced with a binary feature. 
The method then iterates over the unlabelled examples: if an unlabelled example is closer to the unlabelled prototype than the labelled prototype, it is chosen to be a reliable negative example.
Euclidean distance is used in our implementation to calculate the distance between a data point and a prototype.

\setlength{\tabcolsep}{6pt}
\begin{table}[t]
    \centering
    \begin{tabular}{@{}cccccc@{}}
        \toprule
        \textbf{song id} & \textbf{decade} &  \textbf{emotion} &  \textbf{popularity} &  ... \\
        \midrule 
        1 & 2020s & excited & 6 & ... \\
        2 & 2010s & neutral & 4 & ... \\
        3 & 2000s & excited & 5 & ... \\
        \bottomrule
    \end{tabular}
    \caption{Example of items in a leaf, for a music streaming service}
    \label{tab:leaf}
\end{table}

\subsubsection{Learning the Classifier and Extracting the Concept Query:}
Once we have a set of reliable negatives, we fit a binary classifier.
As the concept query can only be in a logical query form, our choice is limited to either a rule based classifier or a classifier which can be used to extract a rule based query.
For our approach, we decide to use a decision tree classifier\cite{quinlan1986induction} because it is trivial to extract logical rules from these models and there are efficient open source implementations for learning decision trees\footnote{\url{https://scikit-learn.org/stable/modules/classes.html}}.
In decision trees, positive instances are typically assigned to multiple leafs unless the tree is of depth 1.
Different positive leafs can be considered as sub-concepts of the target concept.
Hence, the final concept query is the disjunctive combination of the logical query corresponding to each positive leaf.
Also, we want the positive leafs to be pure (containing only positive instances) because queries corresponding to impure leafs would lead to predictions that would contain a lot of false positives, this impacts the quality of the learned concept in terms of customer satisfaction.
There are two possible options for creating a query corresponding to a positive leaf: 1. Use conjunction of each split from root to the selected leaf; 2. Use the instances in the positive leafs to generate the query.  

Extracting the first type of query is straight forward: for each branch that leads to a positive leaf, all the tests at the splits can be combined in a conjunction to form a logical query in the conjunctive normal form, which can then be transformed into the form of definition \ref{def:concept_query}.
The predictions made by final concept query created with this approach is equivalent to the predictions made by the decision tree.
For the second type of query, we use the examples in a positive leaf to generate the most specific query that contains all the examples in the leaf.
To create the most specific query, we use the logical formulation of the query from definition \ref{def:concept_query} and add a literal for each feature value that occurs in the set of examples.
For example, for a music streaming application, assume a leaf contains the examples in table \ref{tab:leaf}.
The most specific query for this leaf would be: $\{(decade = 2000s \lor decade = 2010s \lor decade = 2020s) \land (emotion = excited \lor emotion = neutral) \land (popularity = 4 \lor popularity = 5 \lor popularity = 6) \land ...\}$.
The final concept query is the disjunction of query for each positive leaf.

We refer to the first type as {\em dt-query} - $Q_d = \bigvee_l q_{dl}$, and the second type as {\em items-query} - $Q_s = \bigvee_l q_{sl}$ where $q_{dl}$ and $q_{sl}$ are the queries corresponding to the leaf $l$.
Interesting to note that for a given leaf $l$, $q_{sl}(\mathcal{D}) \subset q_{dl}(\mathcal{D})$.
Using the example from table \ref{tab:leaf}, let's assume that $q_{dl}: (decade = 2020s \lor decade = 2010s \lor decade = 2000s) \land (popularity \neq 7)$, any item (song) $s \in q_{sl}(\mathcal{D})$ will satisfy the query $q_{dl}$ because the values of the features $decade$ and $popularity$ in $q_{sl}(\mathcal{D})$ are limited by the the instances in the leaf which all must satisfy the query $q_{dl}$. Hence, we expect that the $Q_d$ leads to a bigger set of examples compared to $Q_s$,
implying that $Q_d$ will lead to low precision compared to $Q_s$ when compared to the target concept.
We explore this in more detail in section 4.

Another thing to consider is the quality of the positive examples in the set $S$. 
Since $S$ is curated by human experts, it can have noisy data.
There may be many reasons for this: the expert is not experienced enough, or the selection of positive examples is based on a general criterion (e.g., all items from a brand {\em x} should be in $S$).
The concept query based on noisy positives would lead to larger amounts of noise in the items filtered by the query from $\mathcal{D}$, which compromises a good customer experience.
Hence, it is important to be robust against such noisy positives.
For this purpose, instead of generating a query based on each positive leaf, we only do it if the size of the leaf is bigger than a threshold (line 5 in algorithm \ref{algo}).
For this, we introduce a parameter called {\em discard threshold}.
The assumption here is that noisy positives would not be similar to each other or to the positive examples and they would be scattered across the tree in different small positive leafs, discarding those leafs would remove their impact.

\subsection{Problem 2: Learning Concepts from a Collection of Itemsets}
In the following, in the interest of conciseness, we refer to instances and sets of instances as items and itemsets, respectively.
The second problem (definition \ref{def:lp_2}) then boils down to identifying product concepts using a collection of itemsets.
The number of target product concepts is not known.
We solve this problem with a two step approach: first we employ an unsupervised clustering approach to cluster different itemsets that contain `similar' items together, secondly we use each identified cluster to create a product concept based on our approach for the first problem.
For clustering, we use the k-means clustering \cite{teknomo2006k} approach which partitions the data such that each data point is assigned to the cluster with the nearest mean.
K-means clustering uses a distance measure between data-points to assign observation to the clusters.
In our approach, however, as we want to cluster itemsets and not data points, we use a data-point representations of the itemsets which are the input to the clustering algorithm.
We represent each itemset with a centroid, which is calculated based on the binarized version of the data points in the itemset. 
Clustering is performed on these centroids.

\begin{algorithm}[t]
\caption{ConceptLearner}
\hspace*{\algorithmicindent} \textbf{input:} set of itemsets $\mathcal{I}$, all items $\mathbf{\mathcal{D}}$, discard threshold $\mathbf{d}$,
\begin{algorithmic}[1]
\State $Centroids = \{\}$; $Queries = \{\}$
\For{each itemset $P$ in $\mathcal{I}$}
        \State $Centroids$ = $Centroids \cup MakeCentroid(P)$
    \EndFor

\State $Clusters = MakeClusters($Centroids$)$
\For{each cluster $C$ in $Clusters$}
    \State $S = \bigcup_{p \in C}GetOriginalInstances(p)$ 
    \State $Queries = Queries \cup ConceptQueryLearner(S, \mathcal{D}, d)$
    \EndFor

\State \Return $Queries$
\end{algorithmic}
\label{algo:clustering}
\end{algorithm}

Once we have the clusters of centroids, we map the centroids back to the corresponding itemsets.
For each cluster, the itemsets are then combined together in a bigger itemset which makes the input $S$ for the leaning problem \ref{algo}.
We learn a product concept for each of the learned clusters.
Importantly, we expect the clusters generated after the k-means approach to be noisy because we use a distance based approach on a simplified representation of the itemsets. 
Noise robust version of approach for problem 1 can be helpful in dealing with this noise, we study this in section 4.
The pseudo-code is available in algorithm \ref{algo:clustering}.

\section{Experiments}
Our experiments are based on the use case of a streaming company called {\em Tunify}.
We first given an overview of Tunify before explaining our experiments.
The code for the experiments is publicly available\footnote{\url{https://github.com/kgoyal40/automatic-generation-of-product-concepts}}.

\subsection{Use Case: Tunify}
Tunify is a music streaming service for commercial clients, where the customers are businesses looking to play music with a commercial licence.
Business places (like restaurants) often want to play music that fits the ambience of the place.
Tunify provides users with a predefined selection of playlists (called {\em musical contexts}) which contain songs that generate different listening atmospheres (e.g., `Hit Parade' contains hit songs being played on the radio).
These musical contexts are the product concept in our work and from now on we refer to them as such for consistency.
Our two learning problems in the context of Tunify are explained in the introduction.
Tunify maintains a large corpus of songs and defines around 1000 product concepts which are all encoded in database queries using .xml files.
We are provided with the full songs database along with the songs that belong to each product concept.
Tunify uses a set of important attributes to define the concept queries, we use the same set of features for learning.

\begin{figure}[h]
  \centering
  \includegraphics[scale=0.4]{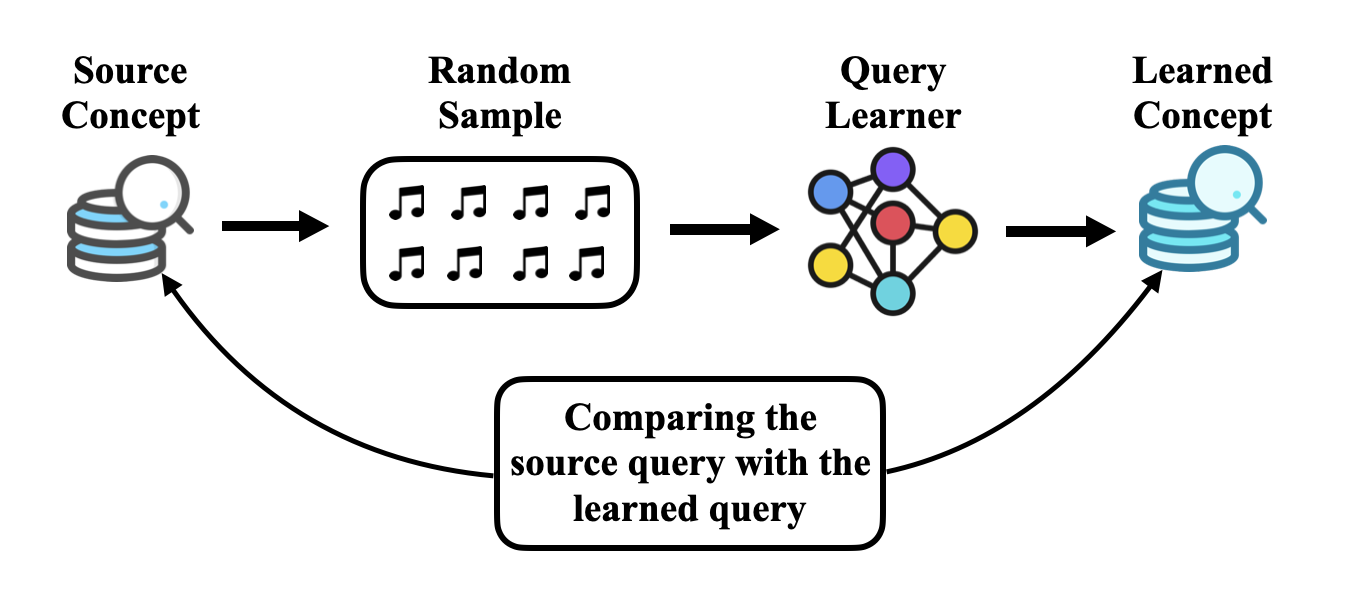}
  \caption{The experimental setup to evaluate the problem 1. We select a concept from the tunify database and sample songs from this concept which becomes our initial set $S$. 
  We learn the query based on this set $S$ and compare it with the source concept.}
  \label{fig:setup_1}
\end{figure}

\subsection{Experiments}
The experimental setup in our case is not straightforward. 
Ideally, for the first learning problem, we would want the music experts to curate a number of initial samples that we can learn concept queries on and then manually evaluate the learned queries and for the second problem, we would like the music experts to manually evaluate all the identified product concepts.
This, however, was infeasible in practice.
Hence, we simulate the experimental setting by using the data provided to us by Tunify.
Next two subsections explain the experimental setup and the evaluation of both learning problems separately.
Experiments were run on an Intel Xeon Gold 6230R CPU@2.10GHz machine with 128 GB RAM.

\subsection{Problem 1: Learning Concept Query from a Set of Items}
For the first learning problem, we aim to answer the following research questions:
\begin{itemize}
    \item {\bf Q1:} Is it possible to learn good quality concept query with our approach?
    \item {\bf Q2:} What is the impact of size of the initial set $S$ on the quality of the learned concept query?
    \item {\bf Q3:} Is the dt-query more general than the items-query?
    \item {\bf Q4:} Do the noisy positives in initial set impact the quality of the learned query negatively? If yes, can we curtail its impact?
\end{itemize}

\subsubsection{Experimental Setup:}
The experimental design for task 1 is detailed in figure \ref{fig:setup_1}.
For this task, we select a total of 184 target product concepts from the Tunify database.
The size $s$ of the initial sample $S$ is varied from 20 to 1000.
After sampling the songs from a target concept $C$, we add some noise by sampling songs from the rest of the database ($\mathcal{D} \setminus Q_C(\mathcal{D})$), where $Q_C(\mathcal{D})$ represents the full set of songs corresponding to concept $C$.
Amount of noise is controlled by a parameter noise ratio ($\mathbf{n}$) which is selected from $[0, 0.1, 0.2]$ in our experiments ($n = 0.1$ means that there are $n*s$ noisy songs in the initial sample), $n = 0$ is the case where there's no noise in the initial sample.
To curtail the effect of the noisy positives, we use the discard threshold ($\mathbf{d}$), which is selected from $[0, 0.1, 0.2]$.
For reliable negatives, we employ two approaches: the likelihood approach which we denote using $`\mathbf{l}'$ and the rocchio approach which we denote using $`\mathbf{r}'$.
We extract two types of queries from the trees: {\em dt-query} and {\em items-query}.
For each target concept and every possible configuration of the parameters, the experiment is repeated 5 times and the average value is reported.

\begin{figure}[h]
\centering
\subfigure[]{\includegraphics[width=.32\textwidth]{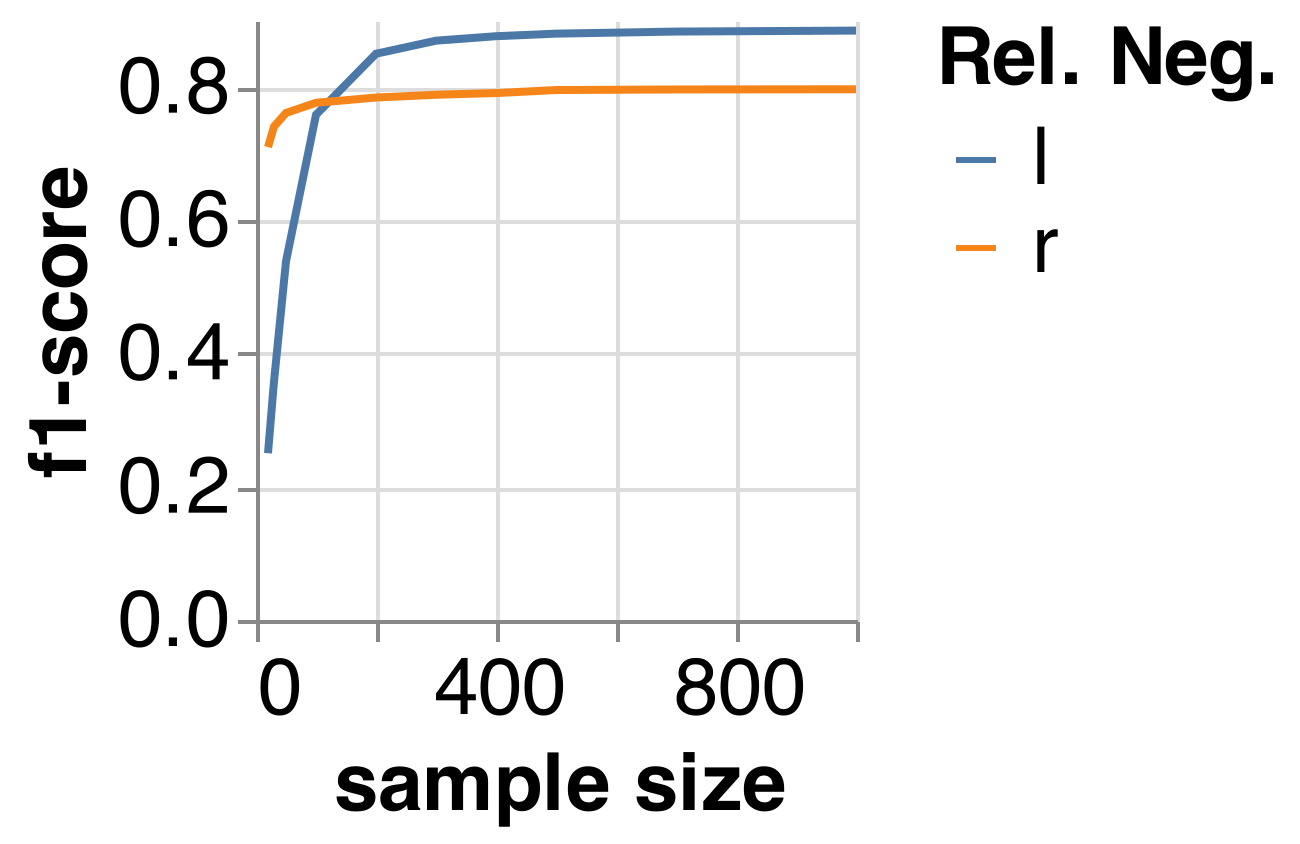}}
\subfigure[]{\includegraphics[width=.32\textwidth]{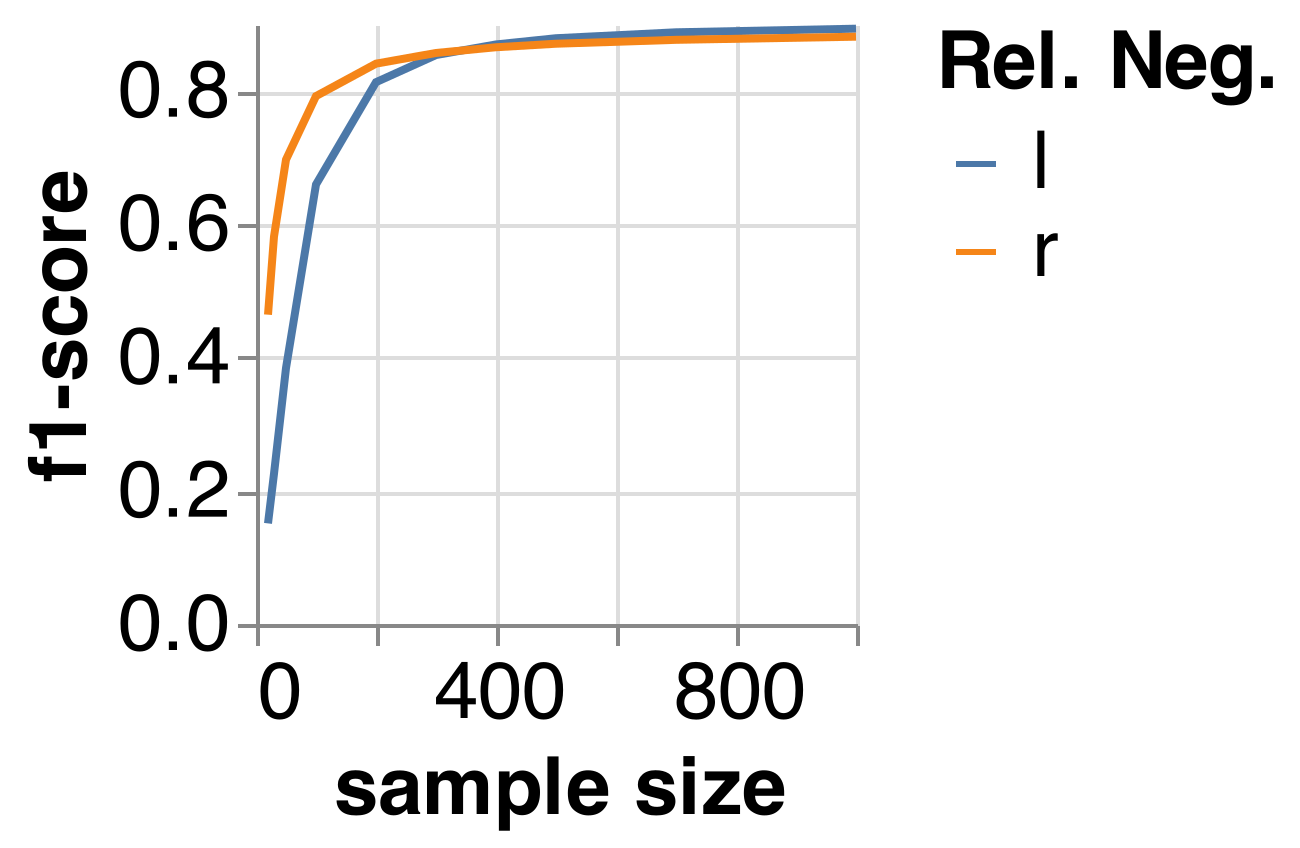}}
\subfigure[]{\includegraphics[width=.32\textwidth]{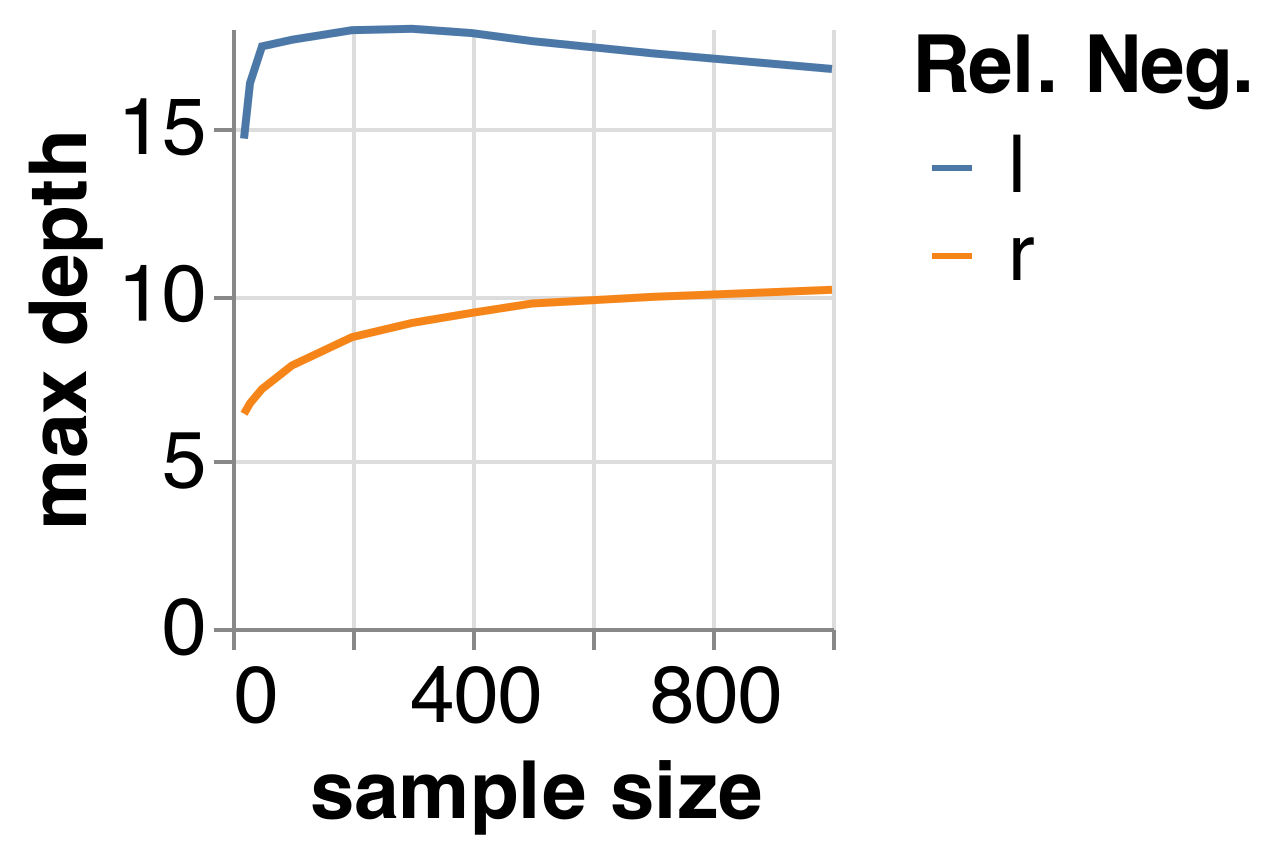}}
\caption{Performance of the query learner when noise ratio = 0. (a) {\bf dt-query}: Rocchio approach performs better than likelihood approach when the number of samples is small but performs worse when the number of samples increases; (b) {\bf items-query}: Rocchio approach consistently performs better than the likelihood approach; (c) the tree learned with rocchio is smaller the tree learned with likelihood approach.}
\label{fig: no noise}
\end{figure}  
\begin{figure}[h]
\centering
\subcapraggedrighttrue
\subcaphangtrue
\subfigure[]{\includegraphics[width=.45\textwidth]{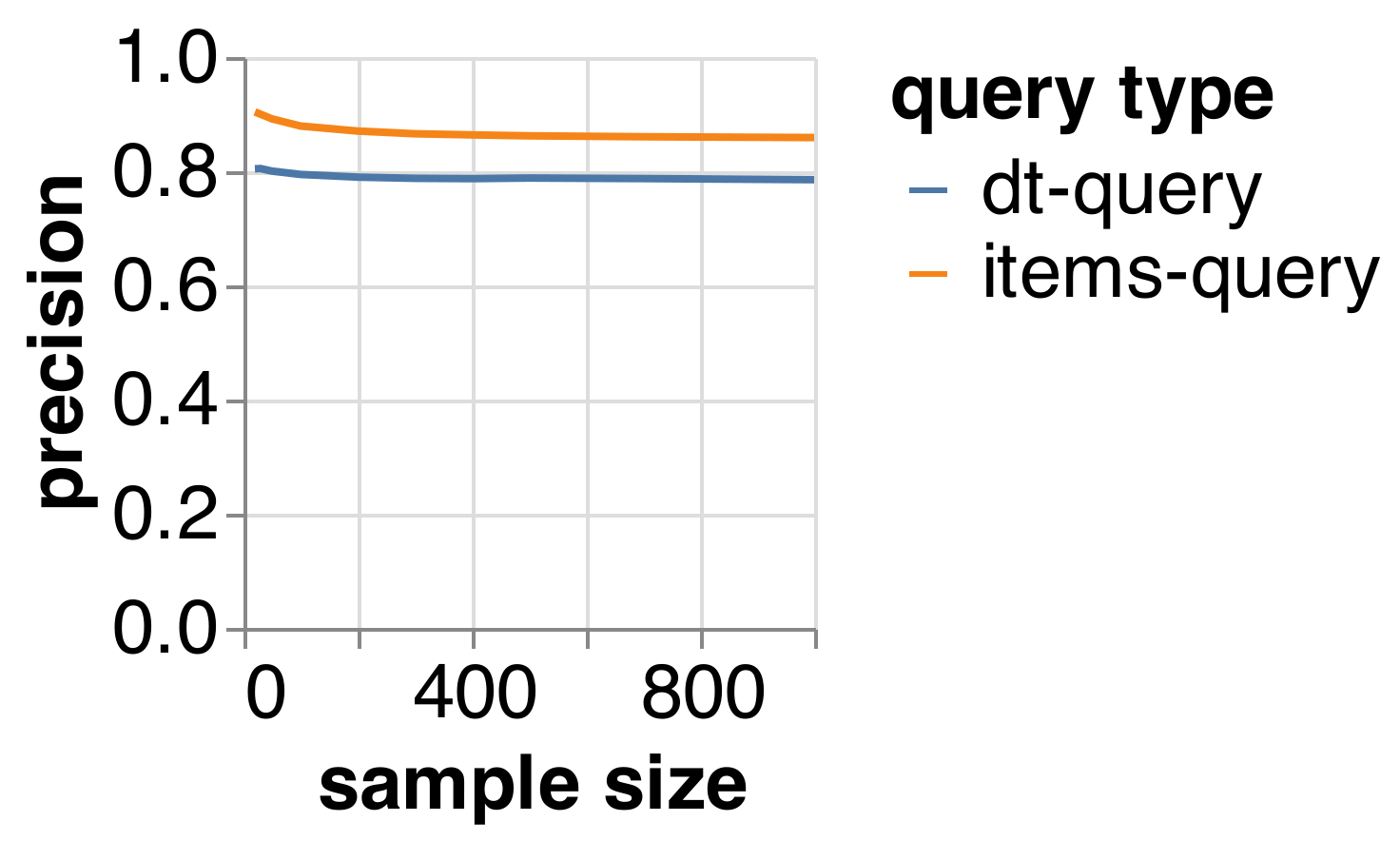}}
\subfigure[]{\includegraphics[width=.45\textwidth]{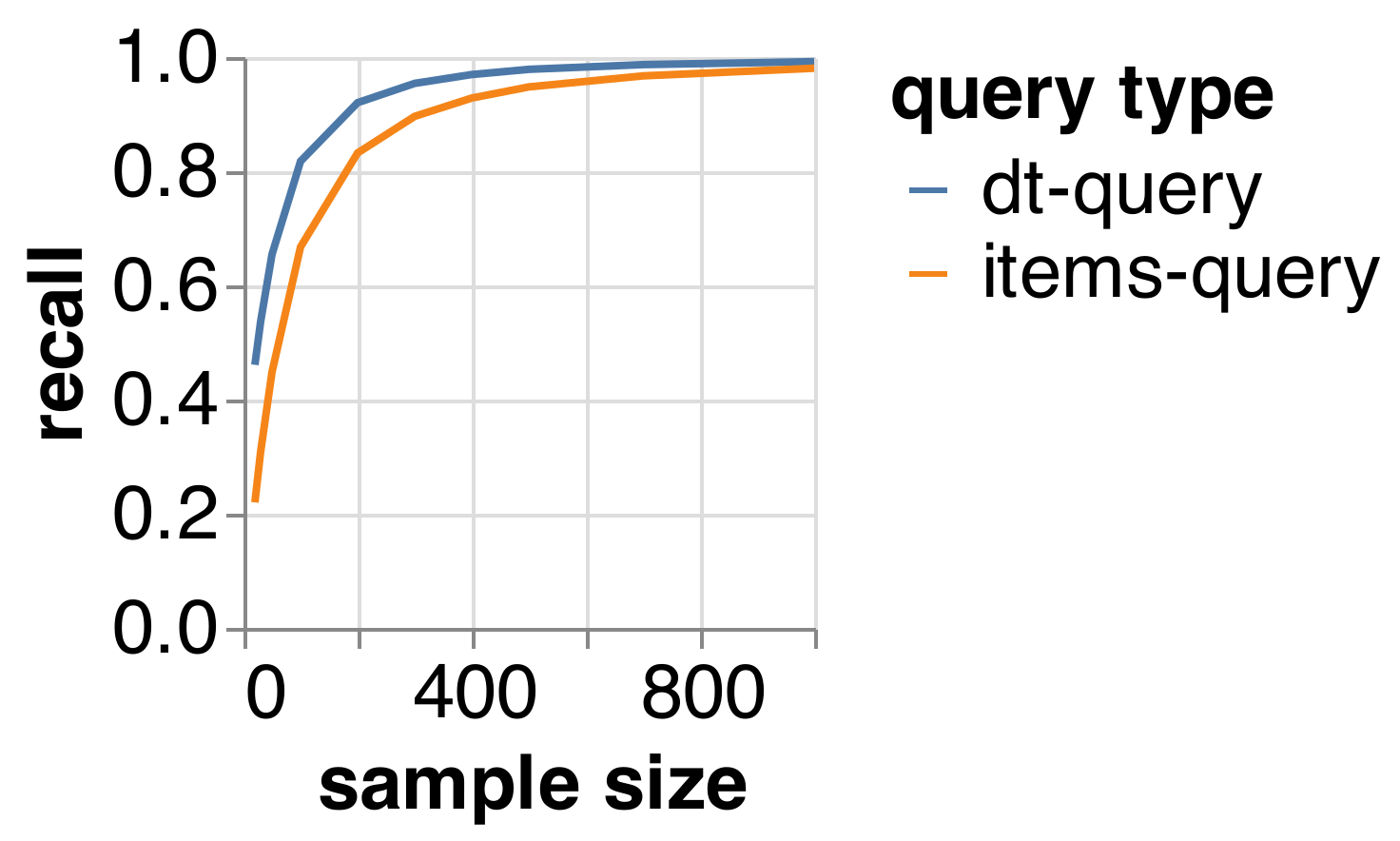}}
\caption{(a) Precision of dt-query is lower than the items-query; (b) Recall for dt-query is higher than the items-query}
\label{fig: prec_recall}
\end{figure} 
\begin{figure}[h]
\begin{center}
\subfigure[$n = 0.1$, $d = 0$]{\includegraphics[width=.32\textwidth]{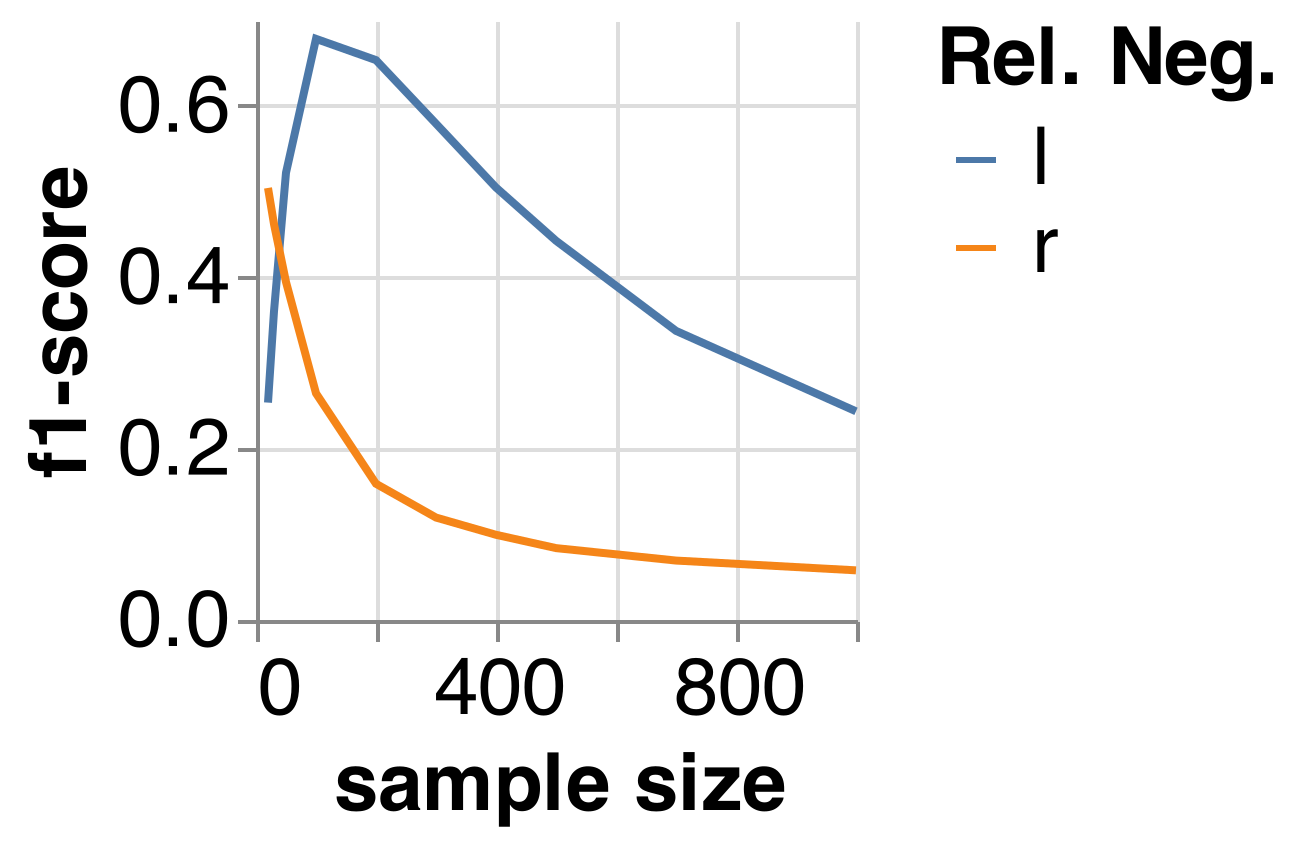}}
\subfigure[$n = 0.1$, $d = 0.1$]{\includegraphics[width=.32\textwidth]{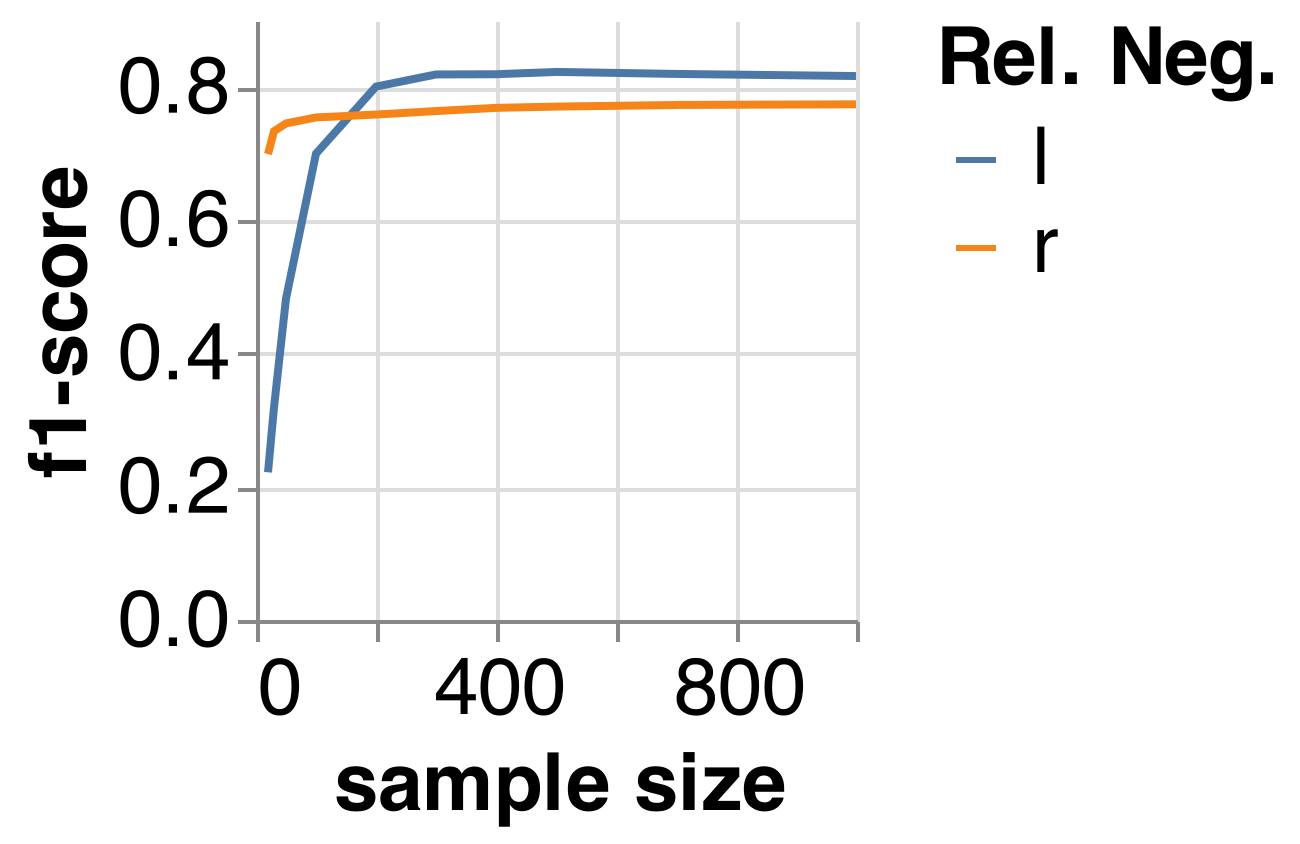}}
\subfigure[$n = 0.1$, $d = 0.2$]{\includegraphics[width=.32\textwidth]{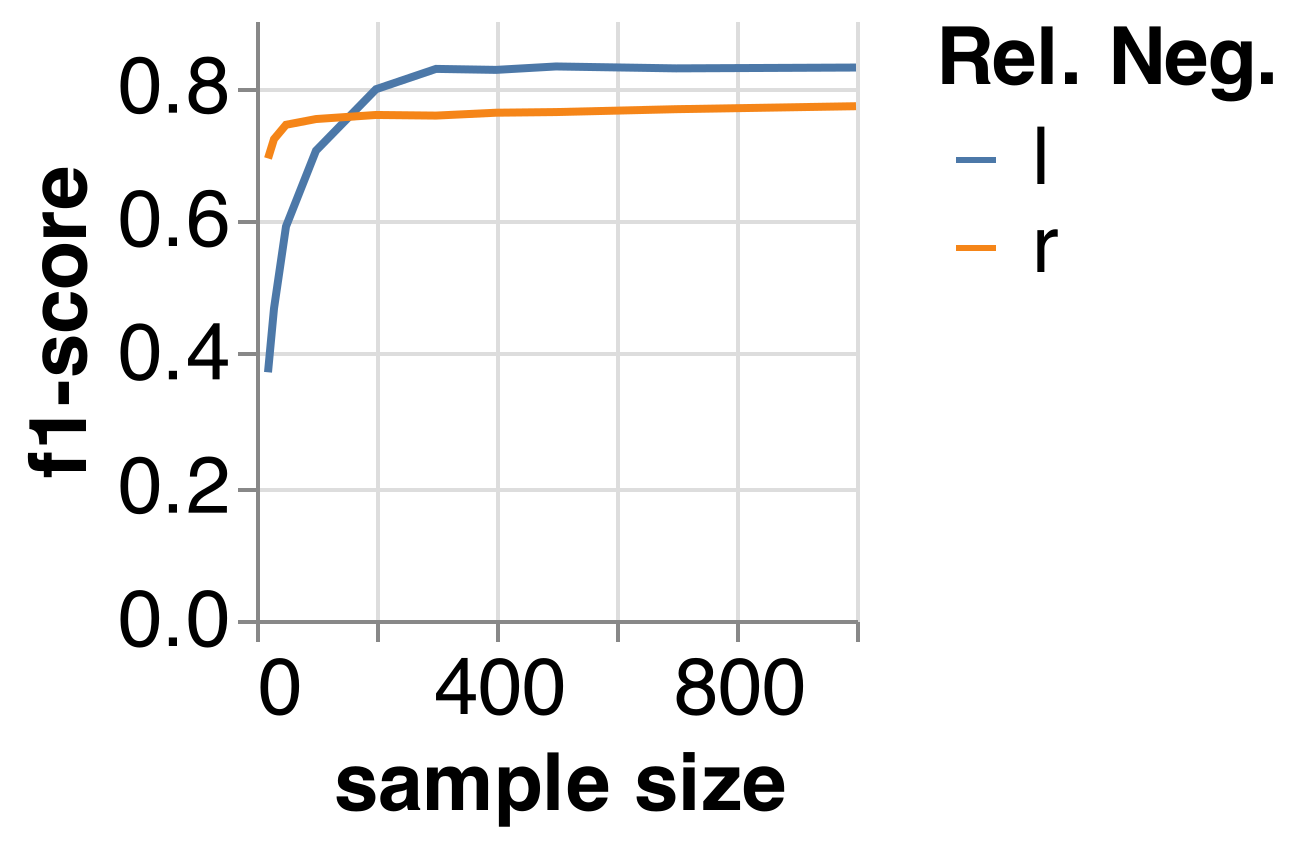}}
\subfigure[$n = 0.2$, $d = 0$]{\includegraphics[width=.32\textwidth]{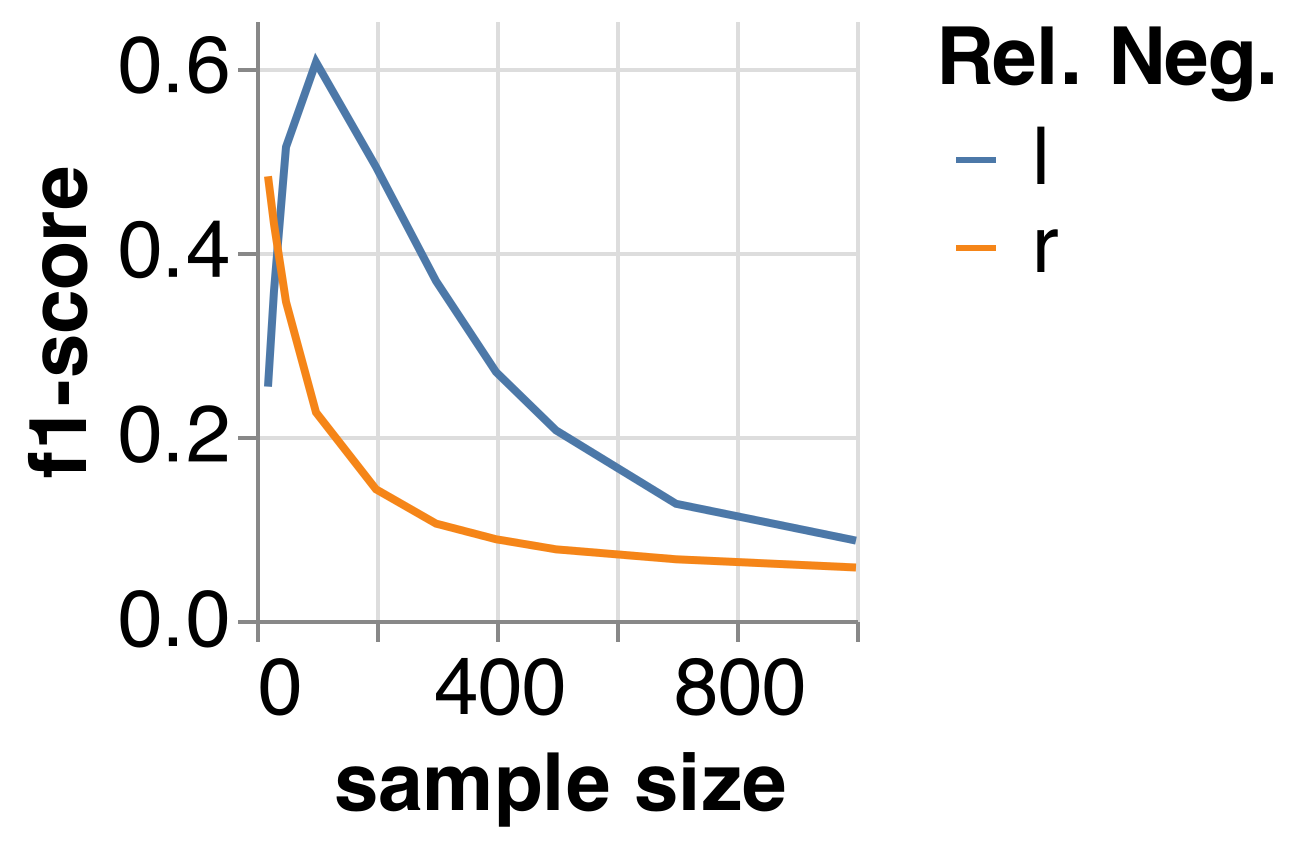}}
\subfigure[$n = 0.2$, $d = 0.1$]{\includegraphics[width=.32\textwidth]{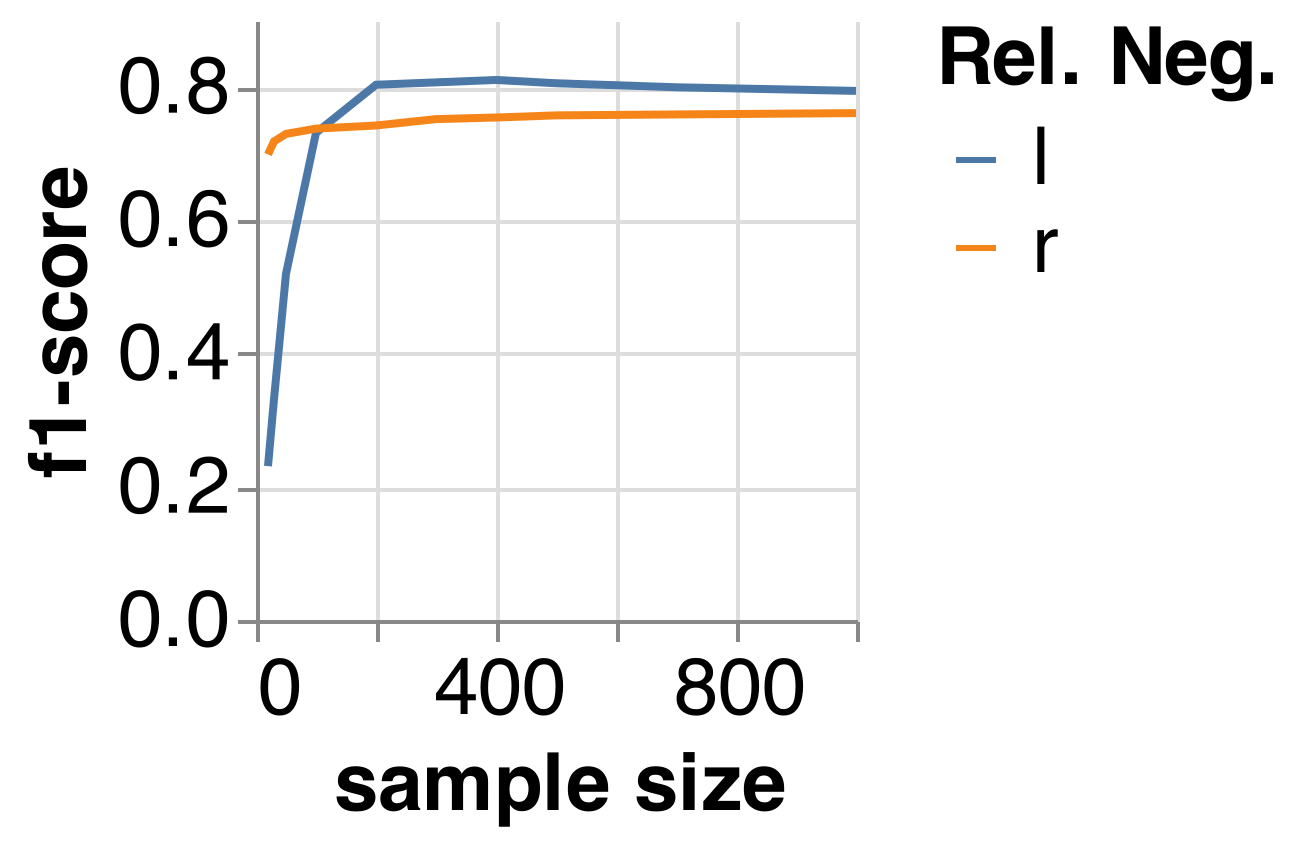}}
\subfigure[$n = 0.2$, $d = 0.2$]{\includegraphics[width=.32\textwidth]{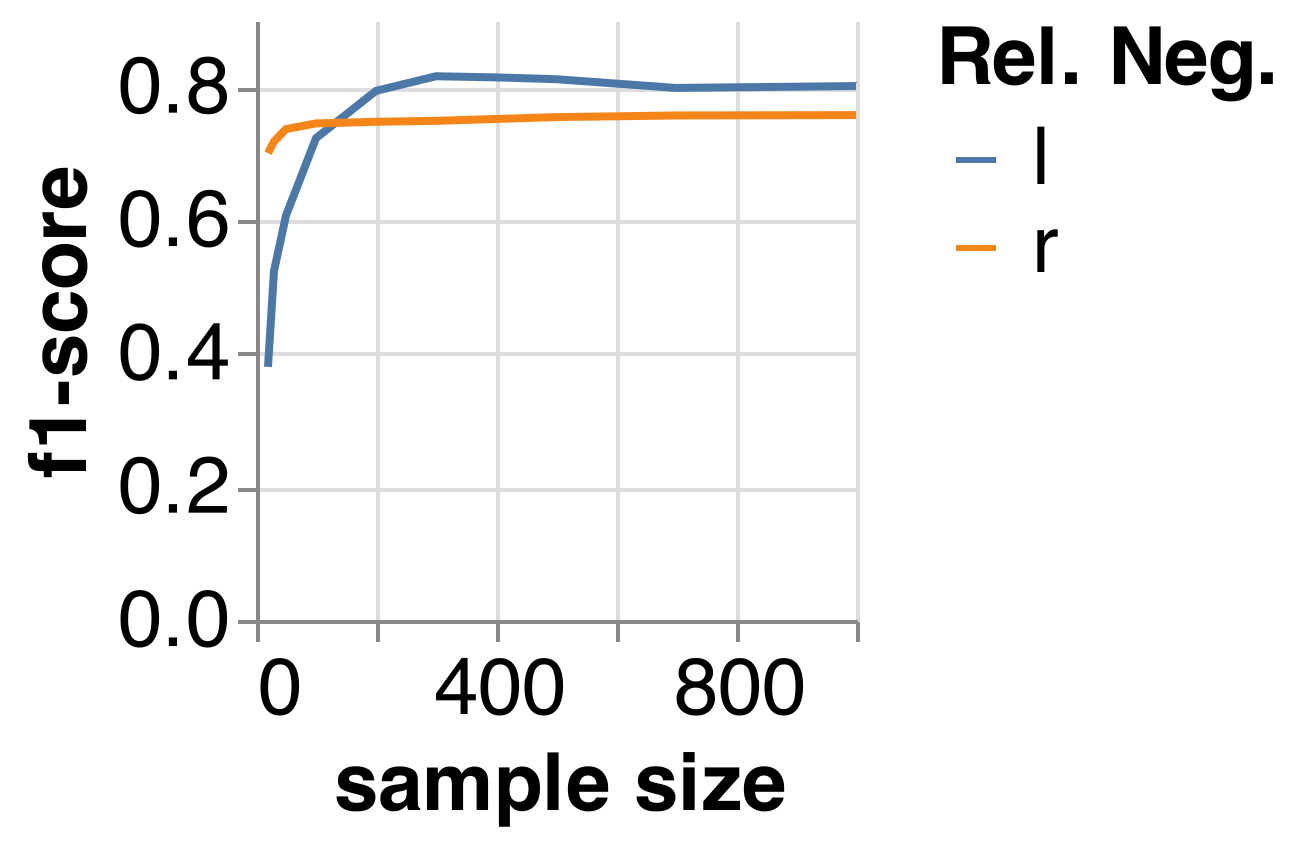}}
\end{center}
\caption{Performance of the query learner for dt-query when noise ratio $n > 0$}
\label{fig: noise dt}
\end{figure}

\begin{figure}[h]
\begin{center}
\subfigure[$n = 0.1$, $d = 0$]{\includegraphics[width=.32\textwidth]{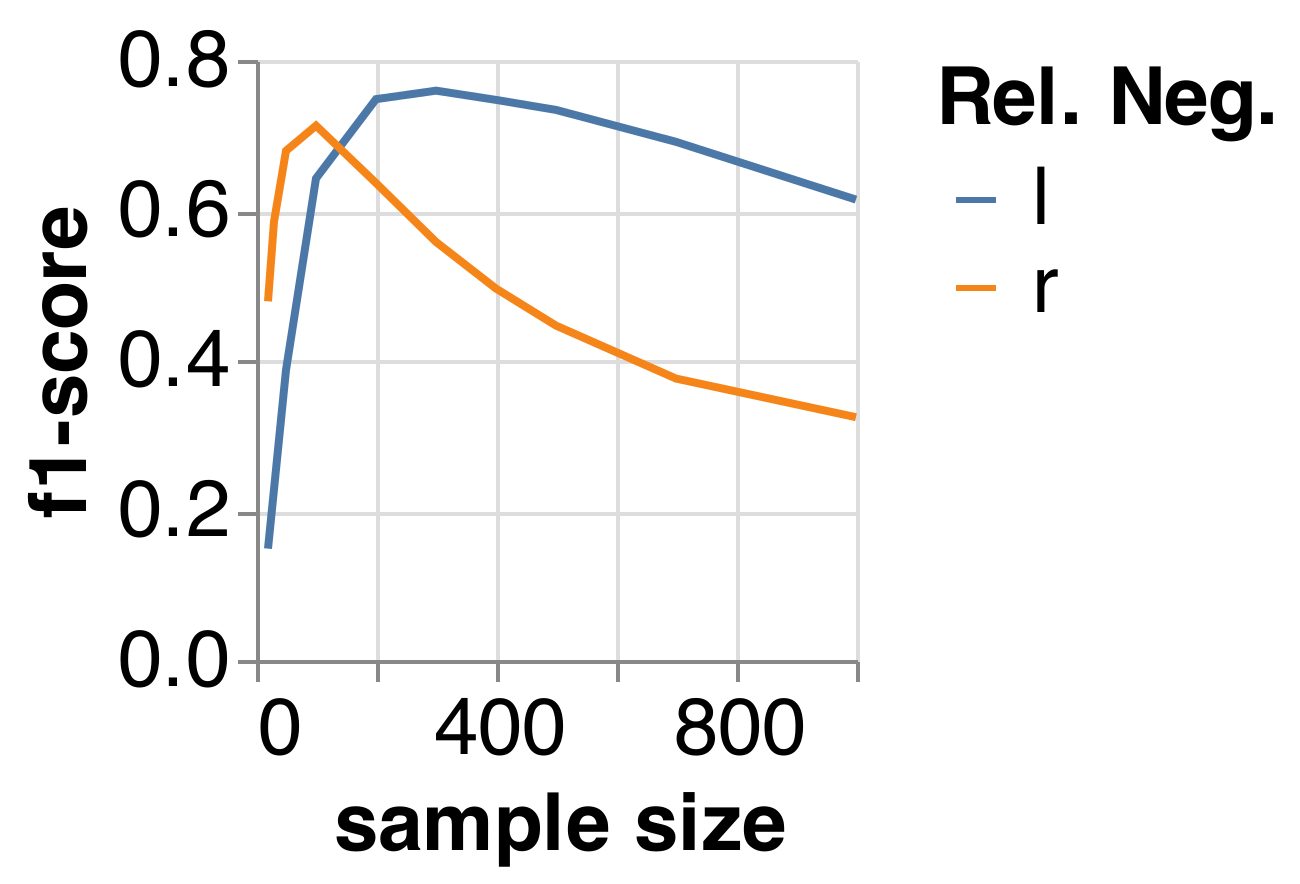}}
\subfigure[$n = 0.1$, $d = 0.1$]{\includegraphics[width=.32\textwidth]{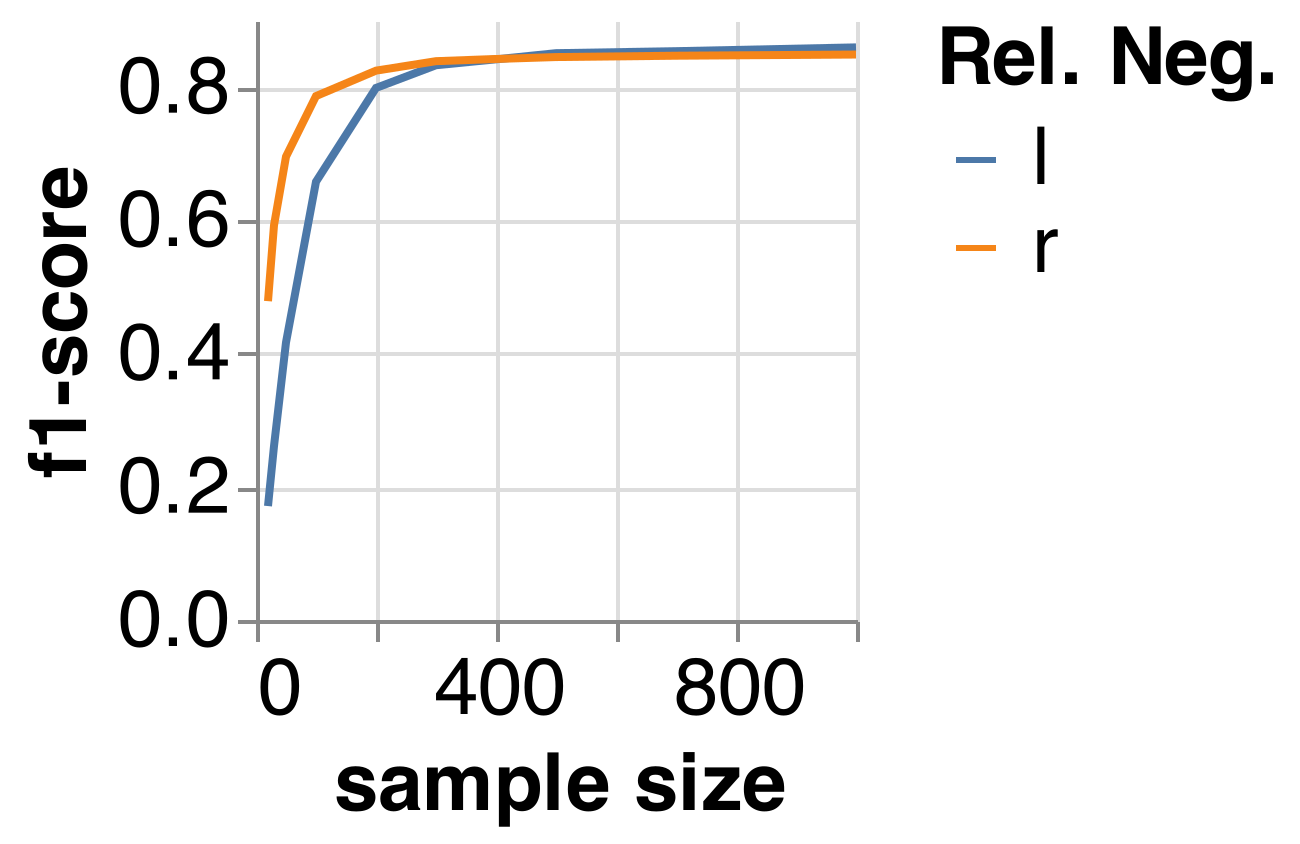}}
\subfigure[$n = 0.1$, $d = 0.2$]{\includegraphics[width=.32\textwidth]{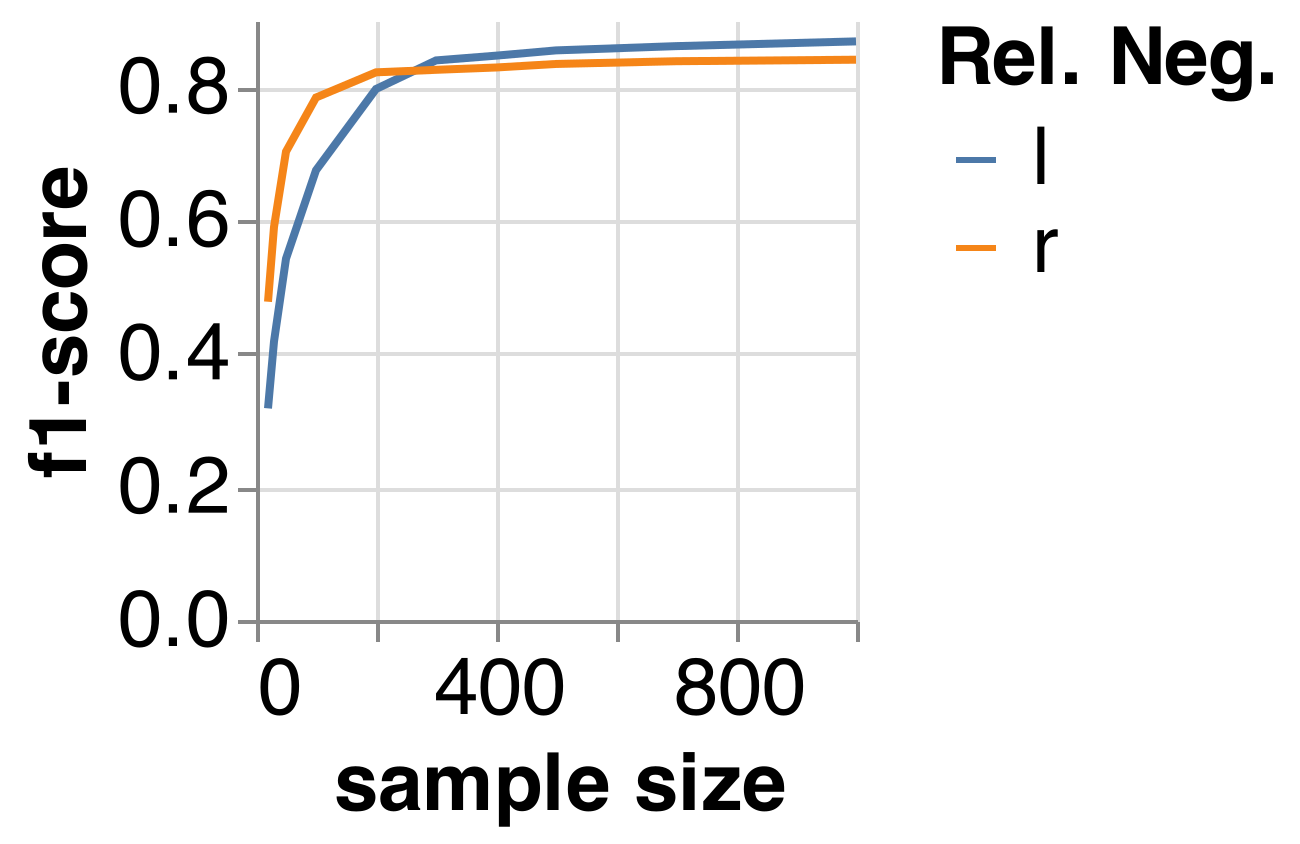}}
\subfigure[$n = 0.2$, $d = 0$]{\includegraphics[width=.32\textwidth]{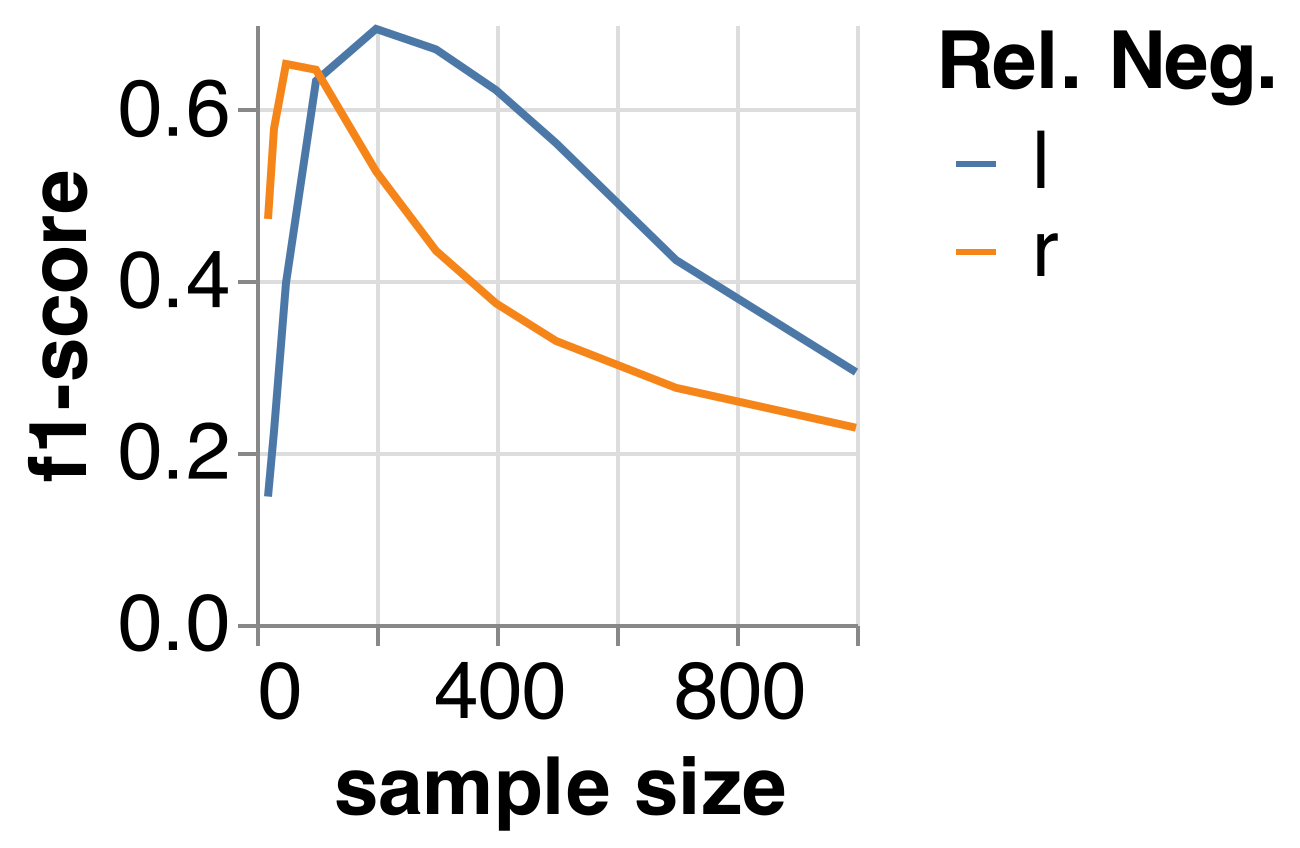}}
\subfigure[$n = 0.2$, $d = 0.1$]{\includegraphics[width=.32\textwidth]{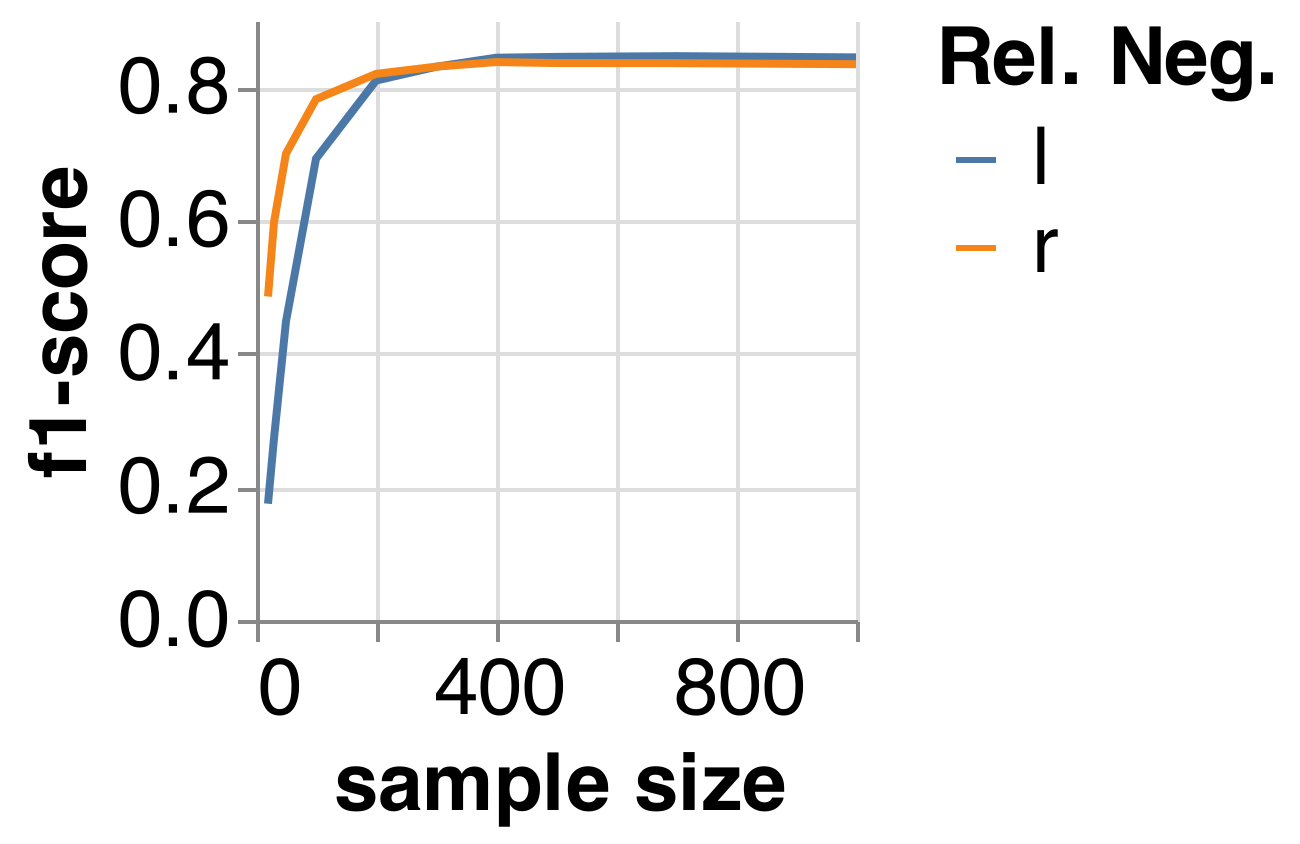}}
\subfigure[$n = 0.2$, $d = 0.2$]{\includegraphics[width=.32\textwidth]{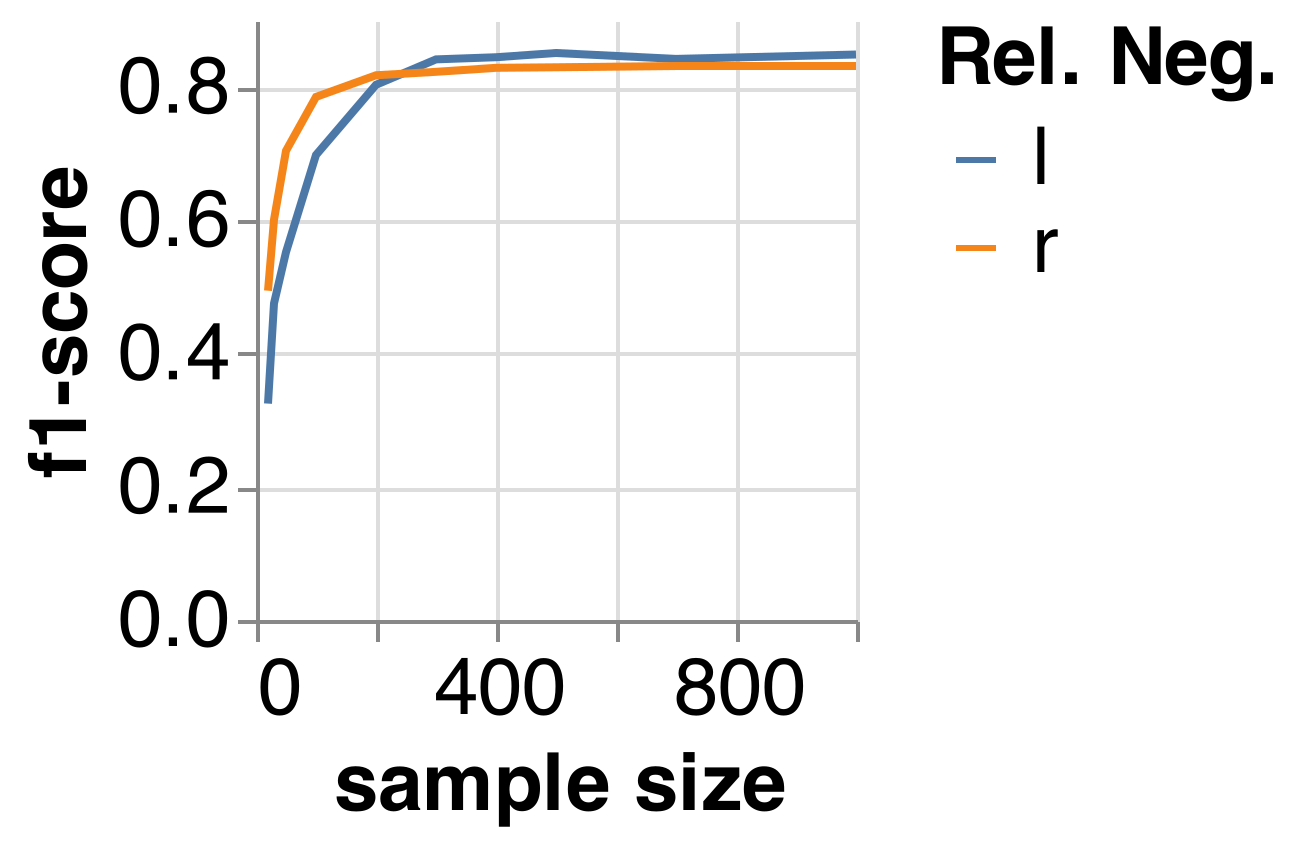}}
\end{center}
\caption{Performance of the query learner for items-query when noise ratio $> 0$}
\label{fig: noise songs}
\end{figure}

\subsubsection{Evaluation:}
We compare the target product concept $C$ with the learned query $Q$ by comparing the sets of songs $Q_C(\mathcal{D})$ with $Q(\mathcal{D})$ using the f1-score\footnote{\url{https://en.wikipedia.org/wiki/F-score}}.
Higher value of f1-score implies higher quality of the learned query $Q$.

\subsubsection{Results:}
\begin{enumerate}
    \item {\bf Learning when there's no noise:} In our base configuration, we select the noise ratio $n = 0$ and discard threshold $d = 0$. Figure 2 represents the f1 score for different sample sizes with different choices for the reliable negative approach, for both items-query and dt-query. 
    Performance of the learned queries improve when the sample size becomes larger before plateauing when size reaches around 200 instances, this answers the research question {\bf Q2}.
    Rocchio is able to learn from smaller number of instances, however it exhibits different behaviour for dt-query and items-query when compared with the likelihood based approach: performance for rocchio drops below the likelihood approach for larger size of initial set, for dt-query.
    This could be explained by the fact that rocchio approach is learning smaller trees (\ref{fig: no noise}(c)) compared to the likelihood based approach, which implies that the size of the dt-query is smaller which implies more generalization and lower precision. 
    The effect would not be there for items-query because they are based on the song instances.
    Additionally, each configuration of reliable negative approach and the query type is able to learn a query with an f1-score of greater than 0.8 even for small sample size, with items-query performing better than the dt-query, this answers the {\bf Q1} positively.
    The reason for items-query performing better than dt-query is because the dt-query generalizes the leaf more than the items-query does, we can check this by looking at precision and recall
    of the dt-query vs items-query in figure \ref{fig: prec_recall}.
    Low precision and high recall for dt-query compared to items-query implies that dt-query predicts more song for a leaf than the corresponding items-query, this positively answers the research question {\bf Q3}.

    \item {\bf Learning with noisy positives:} In the configurations where there are noisy positives, we use the discard thresholds to discard the leafs with small cardinality. 
    Figures \ref{fig: noise dt} and \ref{fig: noise songs} reports f1-scores for different noise ratio $n > 0$ and discard threshold $d$ for dt-query and items-query respectively. 
    As expected, when there are noisy positives we see that the learned queries are not of good quality if the discard threshold is 0: there's a steep decline in f1-score when the data size is increased, this is because more instances lead to more noise which leads to lower precision.
    This behaviour is the same for different query types and reliable negative approaches.
    With non zero discard threshold, our approach is able to treat the impact of the noisy positives on the learned query and we achieve performance similar to the no noise case (f1-score $\geq 0.8)$.
    This answers the research question {\bf Q4} positively.
    An interesting behaviour to note here is that even when the noise ratio is 0.2, best performance is reached at a discard threshold of 0.1, suggesting that more noisy positives doesn't imply bigger noisy leafs.
    
\end{enumerate}

\begin{figure}[h]
  \centering
  \includegraphics[scale=0.4]{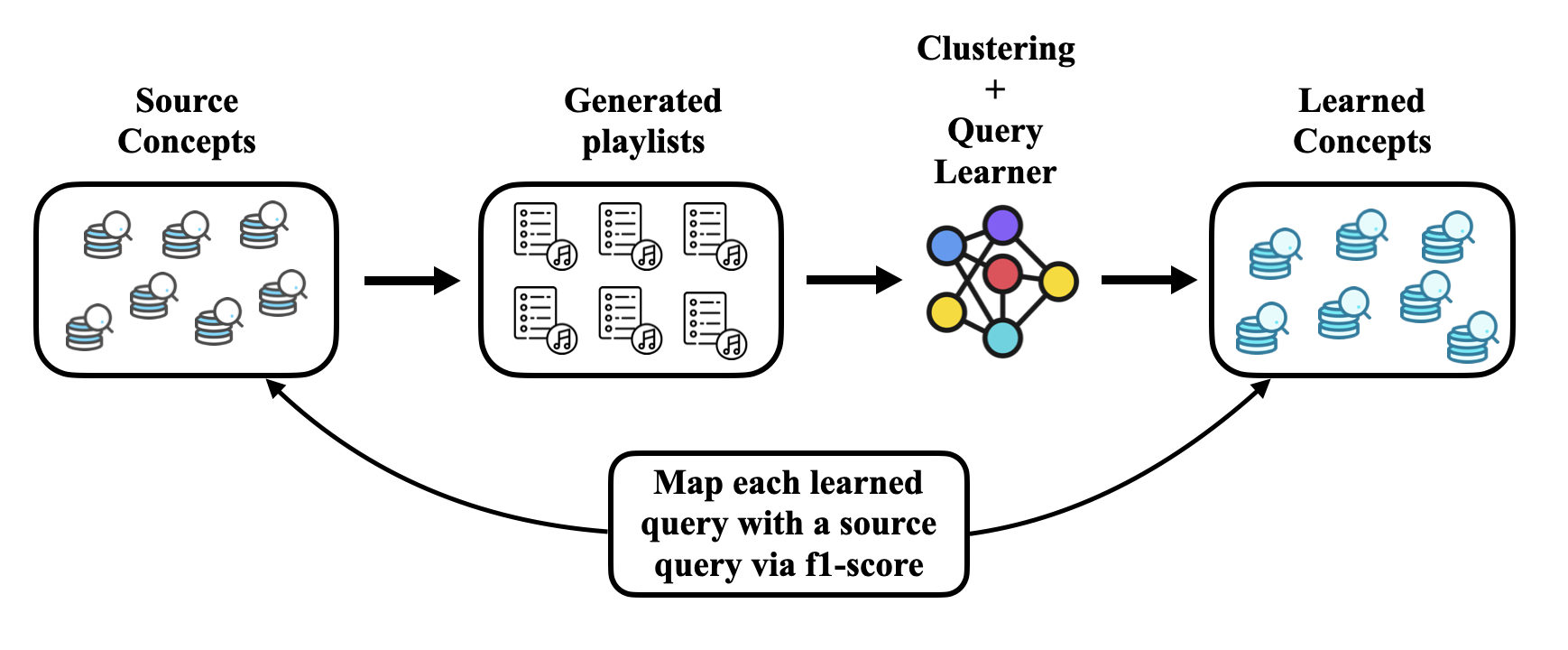}
  \caption{The experimental setup to evaluate the learning problem 2.
  We evaluate with a simulated setup. 
  First, we randomly select a number of product concepts from the database.
  From each of the sampled concepts, we generate a varying number playlists of varying sizes.
  These playlists are then used to generate cluster of songs, which are are used to learn concept queries which are then compared with each of the source concepts.
  The aim is to map a learned concept to one of the source concepts.
  }
  \label{fig:setup_2}
\end{figure}

\begin{figure}[h]
\centering
\subfigure[dt-query]{\includegraphics[width=.4\textwidth]{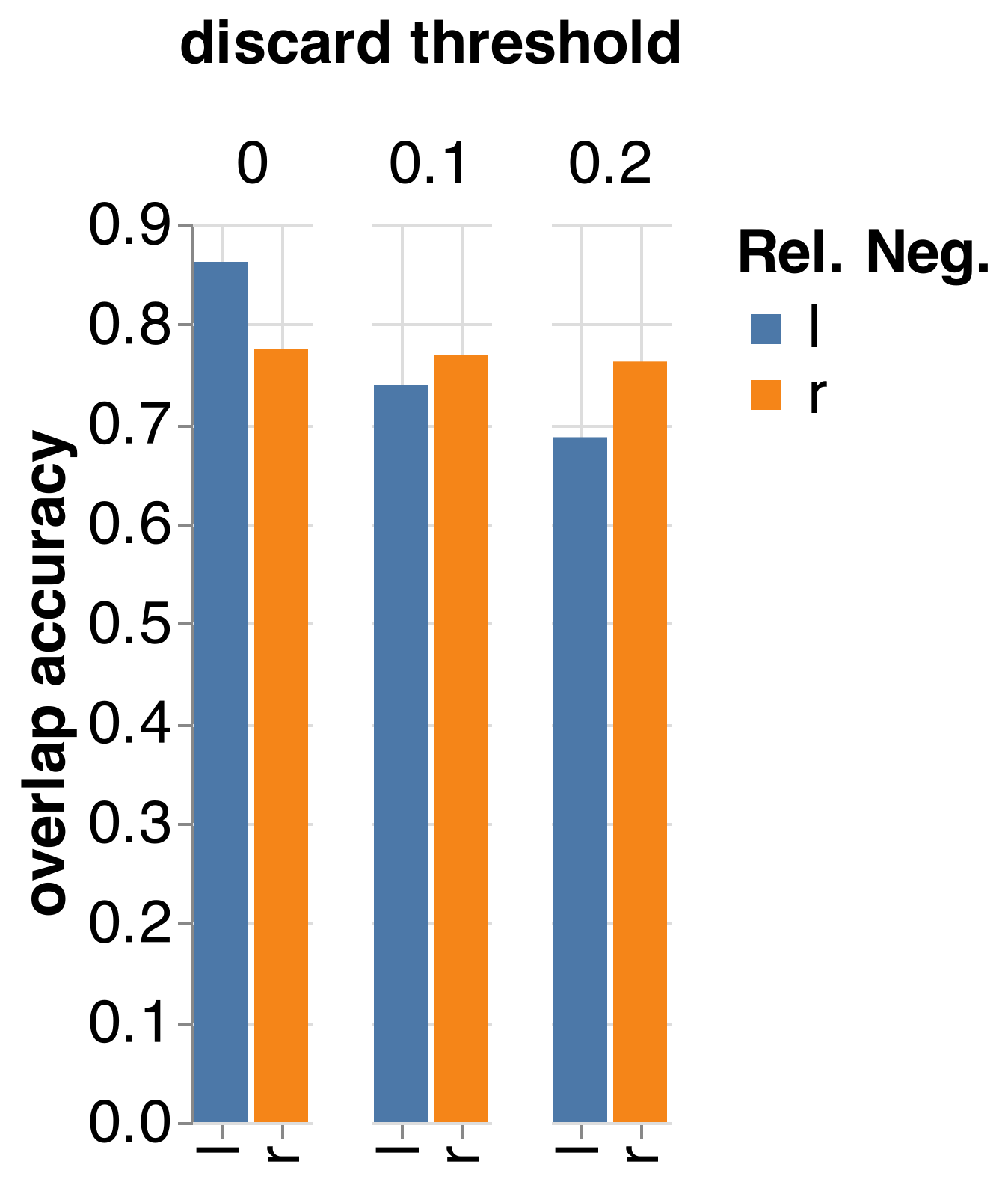}}
\subfigure[items-query]{\includegraphics[width=.4\textwidth]{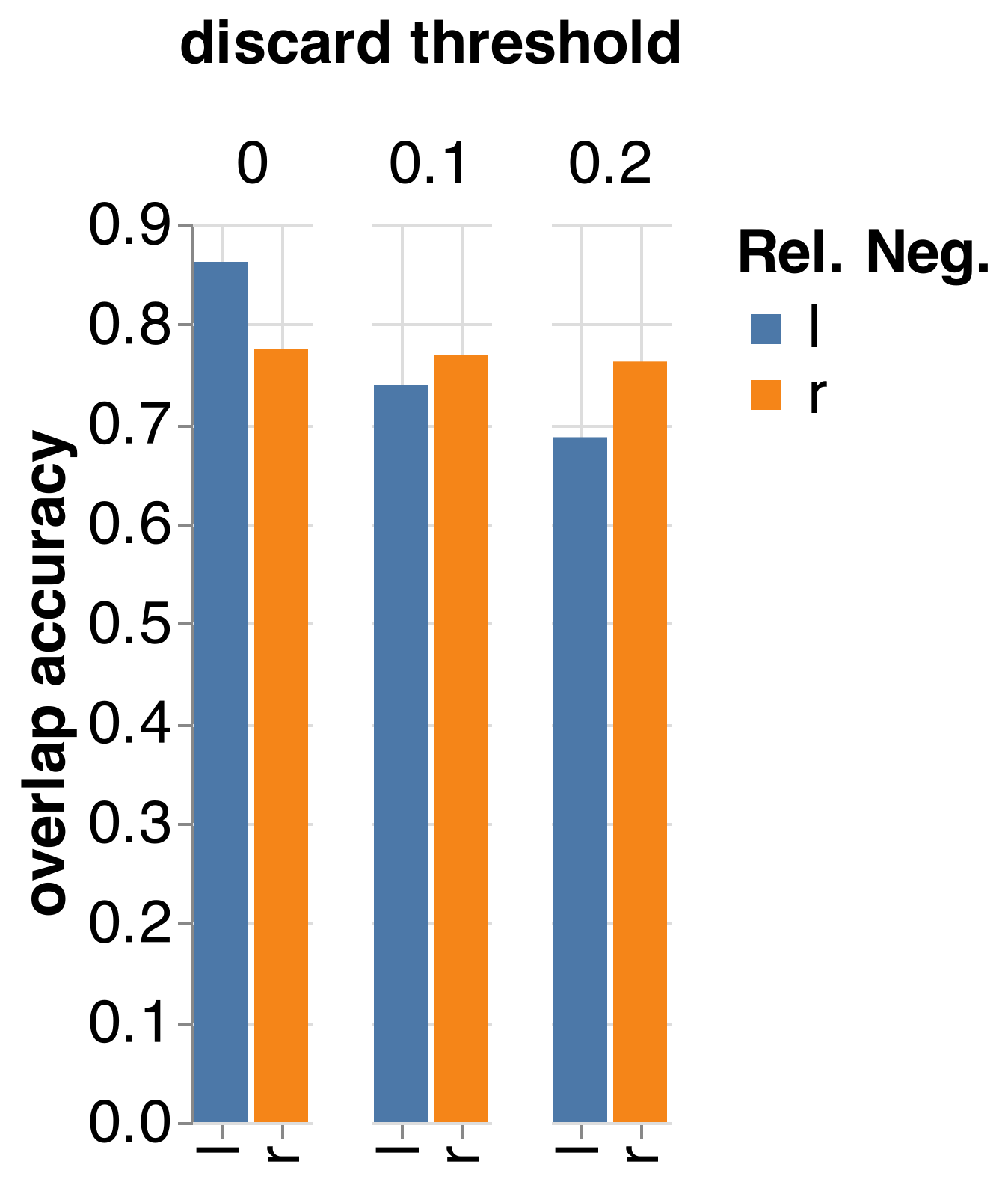}}
\caption{Overlap Accuracy for different configurations of reliable negative approach and discard threshold for both {\bf dt-query} and {\bf items-query}}
\label{fig: clustering}
\end{figure} 

\subsection{Problem 2: Learning Concepts from a Collection of Itemsets}
For the second learning problem, we aim to answer the following questions: {\bf Q1:} Are the learned clusters noisy?; {\bf Q2:} Do the clusters lead to product concepts?


\subsubsection{Experimental Setup:}
Experimental design for task 2 is detailed in figure \ref{fig:setup_2}.
40 source product concepts are randomly selected from the already selected 184 tunify product concepts.
For each selected source concept, somewhere between 5 to 15 playlists are generated randomly.
Each playlist contains somewhere between 15 to 40 songs.
The values discard threshold($d$) is selected from is the same as the previous problem.
We don't induce noise manually because we expect the final learned clusters to be already noisy. 
We repeat the experiments 10 times and the reported values are the average over all the experiments.
For clustering, we use k-means \cite{teknomo2006k} with the number of clusters same as the number of source concepts (40).
In cases where the number of clusters is not known, approaches like the elbow \cite{bholowalia2014ebk} or silhouette distance \cite{lleti2004selecting} can be used to deduce it.

\subsubsection{Evaluation:}
For the evaluation, we use the f1-score to compare the learned concepts and the source concepts. 
We compare a learned concept with all of the source concepts and select the source concept with the highest f1-score. 
The selected source concept is said to be the mapping of the learned concept.
Then for a given configuration, across all experiments, we calculate the percentage of learned concepts where the f1-score of the mapped source concept is greater than $0.7$.
We term this ratio as {\bf `overlap accuracy'}, higher overlap accuracy implies that more learned concepts can be mapped to a source concept.

\subsubsection{Results:}
Results in Figure \ref{fig: clustering} demonstrate that the discard threshold has a negative impact on the performance as we get the best results for $d = 0$, this implies that, the learned clusters are not noisy after all.
This negatively answers the research question {\bf Q1}.
For the dt-query, likelihood approach performs better than rocchio, and for items-query, rocchio performs better than likelihood approach.
This is consistent with our findings in the previous problem.
Overall, we are able to learn a mapping from the learned concept to the source concept in more than $85\%$ of cases for $d = 0$, which positively answers question {\bf Q2}.

\section{Related Work}

The idea of learning playlists automatically using machine learning approaches has been around for many years. In \cite{platt2001learning}, the authors propose a Gaussian regression based approach to generating music playlists using a few seed songs. A probabilistic approach for learning playlists was proposed in \cite{maillet2009steerable}.
There are a number of works that use deep learning approaches for music generation\cite{briot2020deep}. 
In addition to this, machine learning has also been used in various other application involving music data, like music recommendation systems\cite{ayata2018emotion}, automatic music generation \cite{sturm2019machine}, music lyrics generation\cite{malmi2016dopelearning}, automatic music annotation\cite{bahuleyan2018music} and so on.
Based on the application, machine learning can be applied on different aspects of music data like deep learning on audio signals \cite{briot2017deep}, music consumption data for recommender systems \cite{chen2005music}.
In contrast to deep learning models, tree based models have also been explored in music applications \cite{bai2017incorporating,lavner2009decision}.
However, our proposed learning setting, to the best of our knowledge, has not been explored in any of the existing works. 
Also, our framework is general enough to be used in any application where such product concepts are used.

In terms of the approach, we rely on existing methods like rocchio \cite{bekker2020learning} for generating reliable negatives and decision trees \cite{quinlan1986induction} for building the binary classifier.
There's a lot of existing research in PU learning community for generating reliable negatives like 1-dnf \cite{yu2004pebl}, PNLH \cite{fung2005text} and many more \cite{bekker2020learning}.
These methods can also be explored for the reliable negative generation in our approach for concept query generation.
Additionally, for binary classification, other rule based approaches like association rule mining \cite{kotsiantis2006association} can be explored instead of decision trees.
Another important part of our approach is extracting the decision rules from the decision trees.
This approach has been used in many applications in the past for different purposes. Some examples include: using the extracted rules for road safety analysis in \cite{de2013extracting}, for interpretability of neural networks in \cite{938448}.

This paper also relates to the field of constrained machine learning because the syntax of the final model to be used in the system is constrained in a logical form (the product concept is encoded using a logical query in the system).
There are approaches that can learn machine learning models under similar syntactic constraints \cite{goyal2021sade}.
In our work, however, we use decision trees to implicitly enforce such constraints and extract the model in the constrained form (product concept query) to be encoded in Tunify's system.

\section{Discussion}
The dataset provided to us by Tunify is in a tabular form where we assumed independence between features, making it easier for us to calculate the centroid representations of the itemsets for clustering. 
In many real world applications, however, this may not be the case.
Applications might be working with highly relational data, graph data and other complex representations of items. 
Centroid and prototype representations used in the clustering algorithm will be tricky to calculate in such cases.
There is, however, a lot of research in representing relational data using embeddings\cite{narayanan2017graph2vec,perozzi2014deepwalk}.
These techniques could be explored in such scenarios.
Additionally, there are special approaches for clustering in a relational domain, like in \cite{tsitsulin2020graph,long2007relational}, which can be used for our second learning problem.
We leave this as a possible direction for future work.

Another important thing to note here is that our problem setting is very well defined because of the clearly defined product concepts in Tunify. 
The simulated experimental setup makes it relatively easy for us to evaluate the proposed learning problems. 
In real world situations, however, a more thorough qualitative evaluation is needed before such a system can be deployed, either via product experts, or via constant user interaction and feedback using methods like A/B testing\footnote{\url{https://en.wikipedia.org/wiki/A/B_testing}}. 
This was, however, out of the scope of this paper.

\section{Conclusion}
We proposed a decision tree based approach to learning queries defining product concepts from a small number of positive examples.
Additionally, we proposed a clustering based approach to identifying relevant target concepts from a collection of itemsets.
Using our experiments on a real dataset, we demonstrate, for the first task, that the presented approach is able to learn concepts with a high confidence.
Our approach is robust enough to treat the noise in the system and can be used in conjunction to clustering based approaches to identify new product concepts from the data.
For the second task, our simulated experiments show that if the curated itemsets have items that are similar, they will be identified and clustered together in a product concept.
Even though we evaluate on one real world use case of a music streaming service, we believe that this kind of approach can be of interest in various other online businesses where users interact with the systems by using such product concepts.

\section{Acknowledgement}
We thank the reviewers for their constructive input, which helped improve the paper.
This work was jointly supported by the Flanders Innovation \& Entrepreneurship (VLAIO project HBC.2019.2467), Research Foundation - Flanders under EOS No. 30992574, and the Flemish Government (AI Research Program).
We want to thank Tunify for providing us with their data and guidance throughout this project.

\bibliographystyle{splncs04}
\bibliography{main}

\clearpage

\end{document}